\documentclass{article}

\PassOptionsToPackage{numbers, sort&compress}{natbib}

     \usepackage[final]{neurips_2021}

\usepackage[utf8]{inputenc} 
\usepackage[T1]{fontenc}    
\usepackage{hyperref}       
\usepackage{url}            
\usepackage{booktabs}       
\usepackage{amsfonts}       
\usepackage{nicefrac}       
\usepackage{microtype}      

\usepackage{graphicx}
\usepackage{multirow}
\usepackage{amsmath}

\usepackage{subfig}
\usepackage{booktabs}
\usepackage{framed,multirow}
\usepackage{makecell}
\usepackage{placeins}
\usepackage[dvipsnames]{xcolor}
\usepackage{multirow}
\title{A Comparison and Evaluation of Fine-tuned Convolutional Neural Networks to Large Language Models for Image Classification and Segmentation of Brain Tumors on MRI}

\author{
    Felicia Liu\textsuperscript{1},
    Jay J.~Yoo\textsuperscript{2,3,5,7},
    Farzad Khalvati \textsuperscript{2,3,4,5,6,7} 
}

\begin{document}
\maketitle
\newcommand\imgdim{40mm}
\

{\footnotesize \textsuperscript{1}Faculty of Applied Science and Engineering, Engineering Science, University of Toronto, Toronto, ON, Canada\\\textsuperscript{2}Institute of Medical Science, University of Toronto, Toronto, ON, Canada\\
\textsuperscript{3}Department of Diagnostic \& Interventional Radiology, Research Institute, The Hospital for Sick Children, Toronto, ON, Canada\\
\textsuperscript{4}Department of Medical Imaging, University of Toronto, Toronto, ON\\
\textsuperscript{5}Department of Computer Science, University of Toronto, Toronto, ON, Canada\\
\textsuperscript{6}Department of Mechanical and Industrial Engineering, University of Toronto\\
\textsuperscript{7}Vector Institute, Toronto, ON, Canada\\}

\begin{abstract}

Large Language Models (LLMs) have shown strong performance in text-based healthcare tasks. However, their utility in image-based applications remains unexplored. We investigate the effectiveness of LLMs for medical imaging tasks, specifically glioma classification and segmentation, and compare their performance to that of traditional convolutional neural networks (CNNs). Using the BraTS 2020 dataset of multi-modal brain MRIs, we evaluated a general-purpose vision-language LLM (LLaMA 3.2 Instruct) both before and after fine-tuning, and benchmarked its performance against custom 3D CNNs. For glioma classification (Low-Grade vs. High-Grade), the CNN achieved 80\% accuracy and balanced precision and recall. The general LLM reached 76\% accuracy but suffered from a specificity of only 18\%, often misclassifying Low-Grade tumors. Fine-tuning improved specificity to 55\%, but overall performance declined (e.g., accuracy dropped to 72\%). For segmentation, three methods - center point, bounding box, and polygon extraction, were implemented. CNNs accurately localized gliomas, though small tumors were sometimes missed. In contrast, LLMs consistently clustered predictions near the image center, with no distinction of glioma size, location, or placement. Fine-tuning improved output formatting but failed to meaningfully enhance spatial accuracy. The bounding polygon method yielded random, unstructured outputs. Overall, CNNs outperformed LLMs in both tasks. LLMs showed limited spatial understanding and minimal improvement from fine-tuning, indicating that, in their current form, they are not well-suited for image-based tasks. More rigorous fine-tuning or alternative training strategies may be needed for LLMs to achieve better performance, robustness, and utility in the medical space.
\end{abstract}

\section{Introduction}

In recent years, there has been a significant shift in the field of medical artificial intelligence (AI) from traditional convolutional neural networks (CNNs) to large-scale deep learning models, particularly large language models (LLMs). CNNs have demonstrated strong performance in various healthcare applications, including disease diagnosis \cite{tariq_lung_2019}, tumor classification \cite{ayadi2021deep}, and medical image segmentation \cite{ranjbarzadeh_brain_2021}. However, CNNs have inherent limitations, such as their dependence on large annotated datasets and inability to capture long-range, multimodal relationships in medical data. Conversely, LLMs have rapidly evolved and their ability to process and understand large volumes of text has enabled wide adoption in medical natural language processing (NLP) tasks. LLMs have been successfully applied to a variety of healthcare text-based tasks trained using large-scale datasets, transforming healthcare by enhancing clinical decision-making, patient triaging, and automating clinical note generation.

The current applications of LLMs in healthcare primarily assume the availability of large-scale datasets and focus on text-based tasks. This focus overlooks two critical gaps: 1) the performance of LLMs when fine-tuned with limited, domain-specific data, and 2) the applicability of LLMs beyond text, particularly in vision-based tasks such as medical image classification and segmentation. While general-purpose LLMs excel in text processing, there remains a lack of comprehensive overviews evaluating accuracy, robustness, and the overall utility of medical LLMs. When available data is limited, or when applied to non-text-based problems, the effectiveness of medical LLMs is unclear and addressing this gap is central for advancing healthcare AI. This thesis will provide insight into when fine-tuning LLMs is necessary and how they can be adapted for non-textual medical tasks. Understanding how general out-of-the-box LLMs compare with subspecialized models can guide model selection and training strategies for many new medical applications. By broadening their utility, medical LLMs have the potential to revolutionize fields like radiology, leading to more accurate and personalized healthcare.

We evaluated the effectiveness of general-purpose LLMs compared to subspecialized models across various healthcare tasks, with a focus on limited data and non-textual inputs. We first compared general and subspecialized (fine-tuned) models. We assessed whether general LLMs, such as Large Language Model Meta AI (LLaMA) Models, could perform healthcare tasks effectively without fine-tuning, or if specialized training on niche datasets significantly improved performance. We then evaluated LLMs for vision-based tasks by implementing and testing medical tasks such as brain tumor classification and tumor segmentation. Through this research, we hope to provide a more informed understanding of when general models sufficed and when subspecialized models were necessary in healthcare, offering a framework for selecting optimal models and training needs for various medical applications to achieve robust performance.

\section{Materials and methods}
\subsection{Dataset and preprocessing}

\subsubsection{Dataset}
We used the (Brain Tumor Segmentation) dataset consisting of multi-modal MRI scans with T1-weighted (T1), T1-weighted with contrast enhancement (T1ce), T2-weighted (T2), and Fluid-Attenuated Inversion Recovery (FLAIR) images. These imaging modalities together give a comprehensive view of the tumor’s structure and its interaction with surrounding tissues. Additionally, the dataset included expert-annotated segmentation masks and each patient was labeled as either an HGG or LGG patient. The BraTS 2020 dataset was chosen due to its relevance and quality for glioma classification, its annotations, and its relatively larger sample size of 365 patients make it an ideal dataset for training and testing models \cite{bakas_identifying_2019, bakas_advancing_2017, menze_multimodal_2015, bakas_dataset, noauthor_brats_lgg}.

\subsubsection{Classification data processing}
\begin{figure}[h!]
\centering
\includegraphics[width=0.8\textwidth]{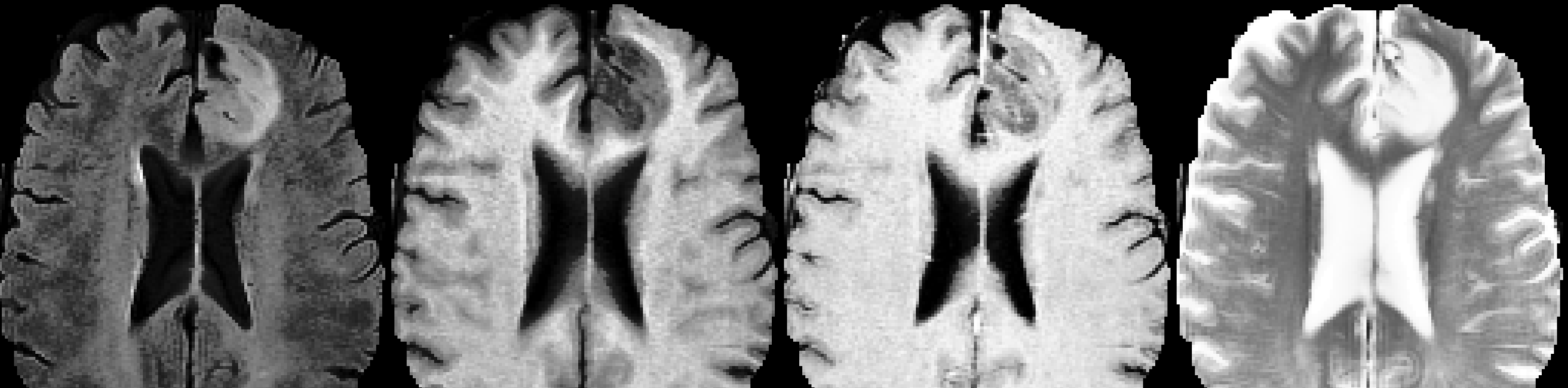}
\caption{Scan modality visualization for an LGG patient \cite{bakas_identifying_2019, bakas_advancing_2017, menze_multimodal_2015, bakas_dataset, noauthor_brats_lgg}.}
\label{fig7}
\end{figure}

\begin{figure}[h!]
\centering
\includegraphics[width=0.8\textwidth]{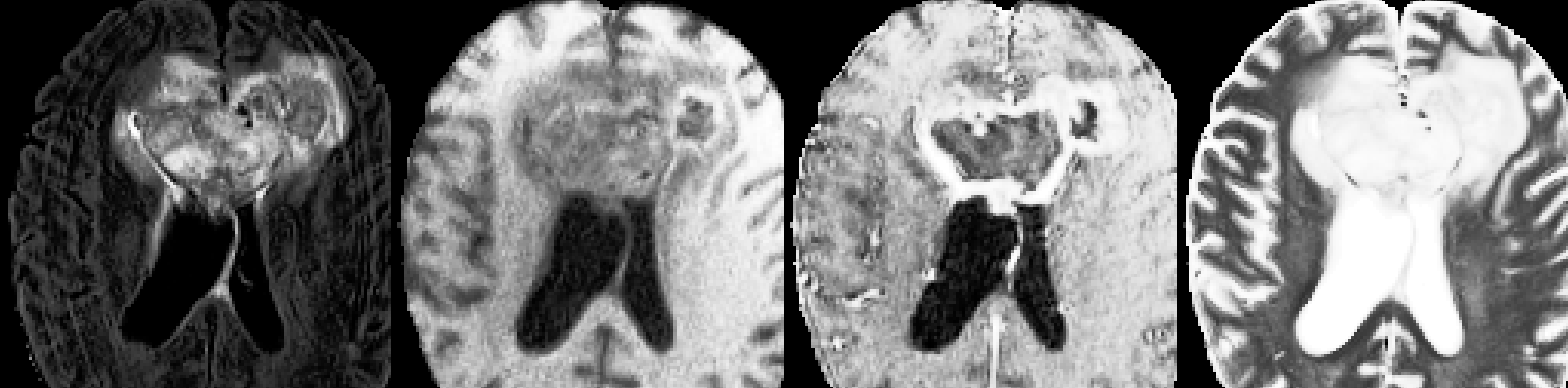}
\caption{Scan modality visualization for an HGG patient \cite{bakas_identifying_2019, bakas_advancing_2017, menze_multimodal_2015, bakas_dataset, noauthor_brats_lgg}.}
\label{fig8}
\end{figure}

All experiments, including classification and segmentation tasks using CNNs and LLMs, were conducted with the same cleaned data to ensure a fair comparison. The BraTS 2020 dataset consists of 365 patients, each with four MRI scans (T1, T1ce, T2, and FLAIR) providing 95 axial slices per scan, resulting in 95x128x128 voxel arrays of intensity. The scans were processed as integer arrays (0-256 per pixel), normalized, and centered around the glioma using radiologist-annotated segmentations. All images were resized to a consistent shape of (4, 95, 128, 128). To focus on the capabilities of the CNN and LLM models, no additional image processing techniques, such as filtering or histogram equalization, were applied. The dataset was split into separate cohorts: 310 training, 62 validation (for each of 5-folds), and 55 test samples (15\%). Care was taken to ensure that all scans and slices for a single patient were kept within the same cohort, preventing any overlap of information between the training, validation, and test sets. The dataset was also imbalanced in that there are far fewer LGG patients compared to the HGG patients. All training and validation data was balanced by oversampling the LGG class. Sample patient scans from the BraTS 2020 dataset are given in Figure~\ref{fig7} and Figure~\ref{fig8}. 

The BraTS 2020 dataset labels each patient as either LGG or HGG, using 0 for LGG and 1 for HGG. These labels, stored alongside the MRI images, were extracted and translated into binary format to train models for distinguishing between low and high-grade gliomas used in the CNN classification tasks.

\subsubsection{Segmentation data processing}
For the segmentation tasks, each of the 365 patients had a 95x128x128 scan structured as a voxel occupancy grid, where zero values indicated regions without glioma and non-zero values represented glioma presence. To ensure viable segmentation, the first and last slices containing glioma were identified in the axial, coronal, and sagittal planes, allowing for the exclusion of non-glioma slices, ensuring that only relevant slices were used in training, thus minimizing potential confusion. Three types of segmentation data were extracted. The first method involved defining a bounding box for each axial slice containing glioma. This was done by extracting the minimum and maximum coordinates along both the x and y axes (xmin, xmax, ymin, ymax), creating a square bounding box. The bounding box was represented by four points: top-left, top-right, bottom-right, and bottom-left, stored in a (4, 2) array for each slice. The second approach focused on the center of the glioma, where the center of the bounding box was calculated and recorded as a single (1, 2) coordinate for each axial slice containing glioma. This representation was simpler but still provided the important information on glioma localization. The third, most complex approach involved representing the glioma with polygons. For each axial slice, the boundary of the glioma was traced using 10 to 15 points, following a clockwise direction starting from the north-most point. Smaller gliomas used fewer points, while larger or more irregularly shaped ones used more. The polygon data was stored as a list of coordinates with shape (10-15, 2), capturing the detailed contours of the glioma. This method offered the most precise and detailed segmentation, providing data for training segmentation models with a higher degree of accuracy in tumor shape and location.

\begin{figure}[h!]
\centering
\includegraphics[width=1.0\textwidth]{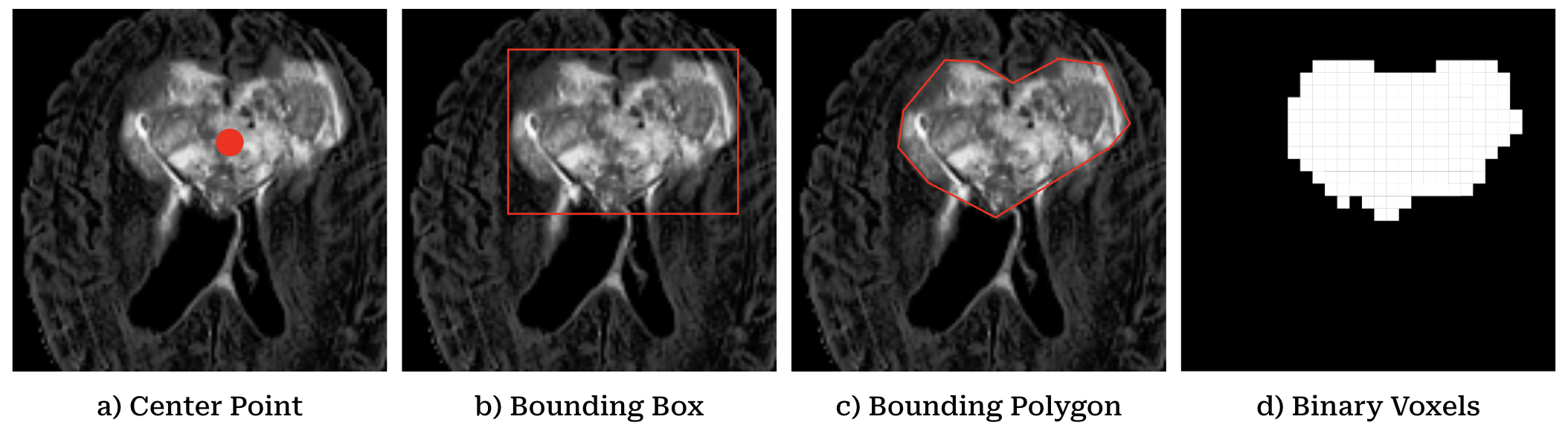}
\caption{Sample segmentations using a) center points, b) bounding boxes, c) bounding polygons, and d) the given form of binary voxels.}
\label{fig82}
\end{figure}

\subsubsection{Creating conversation datasets for fine-tuning}
For all classification and segmentation tasks, the data was structured into a format compatible with Unsloth for LLM fine-tuning, converting raw information into a conversation dataset. This format was consistent across tasks, with slight variations depending on the specific instructions and ground truth data. The conversation structure included the task instruction and relevant image data, ensuring the LLM had the necessary context. Each task began with a clear instruction (e.g., classifying brain scans or segmenting glioma regions) embedded in the dataset. For classification tasks, the label (e.g., “Low Grade Glioma”, “High Grade Glioma”) served as the ground truth, while segmentation tasks used center points, bounding boxes, or polygons. The conversation dataset was organized as exchanges between a "user" and an "assistant," with the assistant’s response based on the ground truth data. This approach ensured proper preparation for fine-tuning the LLM, making it easy to feed into Unsloth and adapt the model for specific tasks. The process was standardized for this efficient training while preserving task-specific details.

\begin{figure}[h!]
\centering
\includegraphics[width=0.9\textwidth]{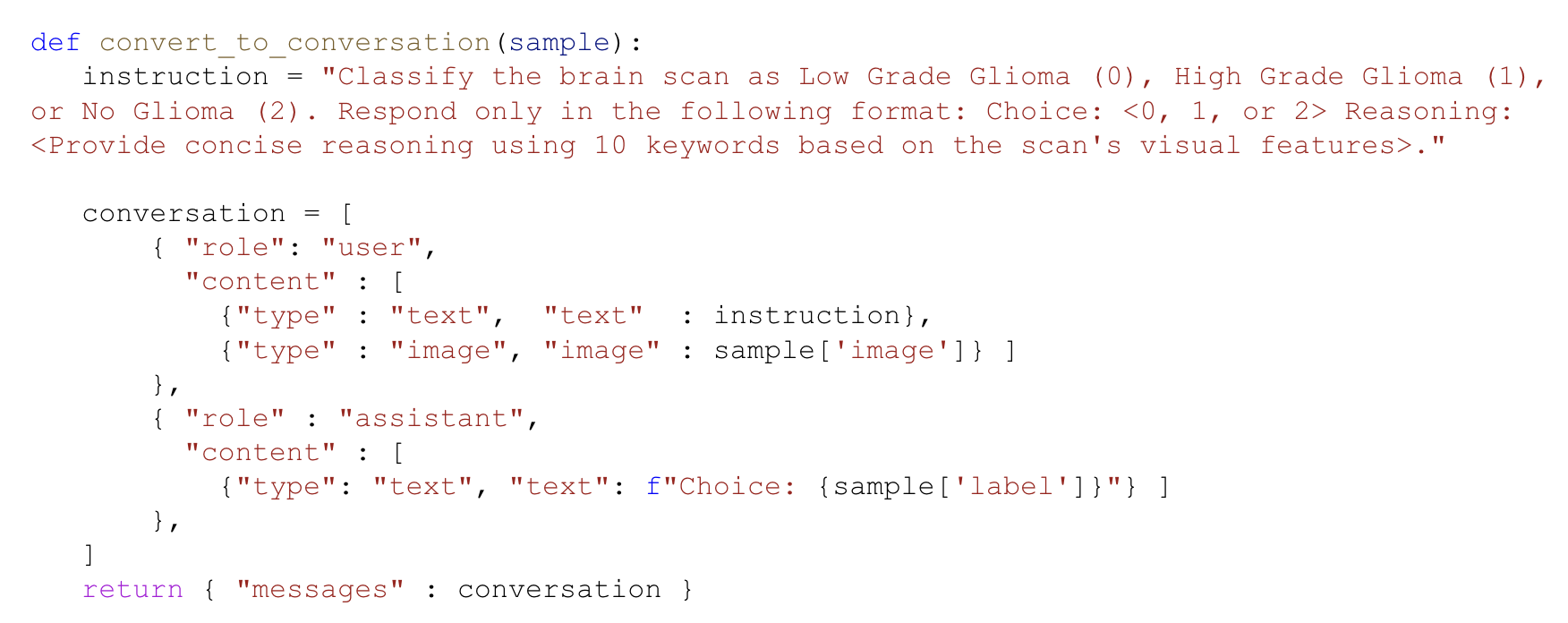}
\caption{Sample LLM conversation dataset conversion for the classification task.}
\label{fig83}
\end{figure}

\subsection{Methodology}

\subsubsection{Model inputs}

Due to differences in input processing, the CNN and LLM approaches differed in how the data was processed for training, inference, and evaluation. The CNN model leveraged full 3D convolutions across all four imaging modalities (T1, T1ce, T2, FLAIR), enabling it to capture comprehensive spatial information from the entire volume of each scan. With the power of GPUs, which process the data quickly, the full 3D scans for each patient were fed into the model, rather than processing each 2D slice individually. This allowed the CNN to discern a single metric per patient, such as one LGG/HGG classification or one output voxel segmentation mask, which directly corresponds to the 3D ground truth mask for that patient. This approach benefits from the ability to process the entire scan, leveraging spatial relationships between slices, offering a more holistic view of the tumor. 

\begin{figure}[h!]
\centering
\includegraphics[width=1.0\textwidth]{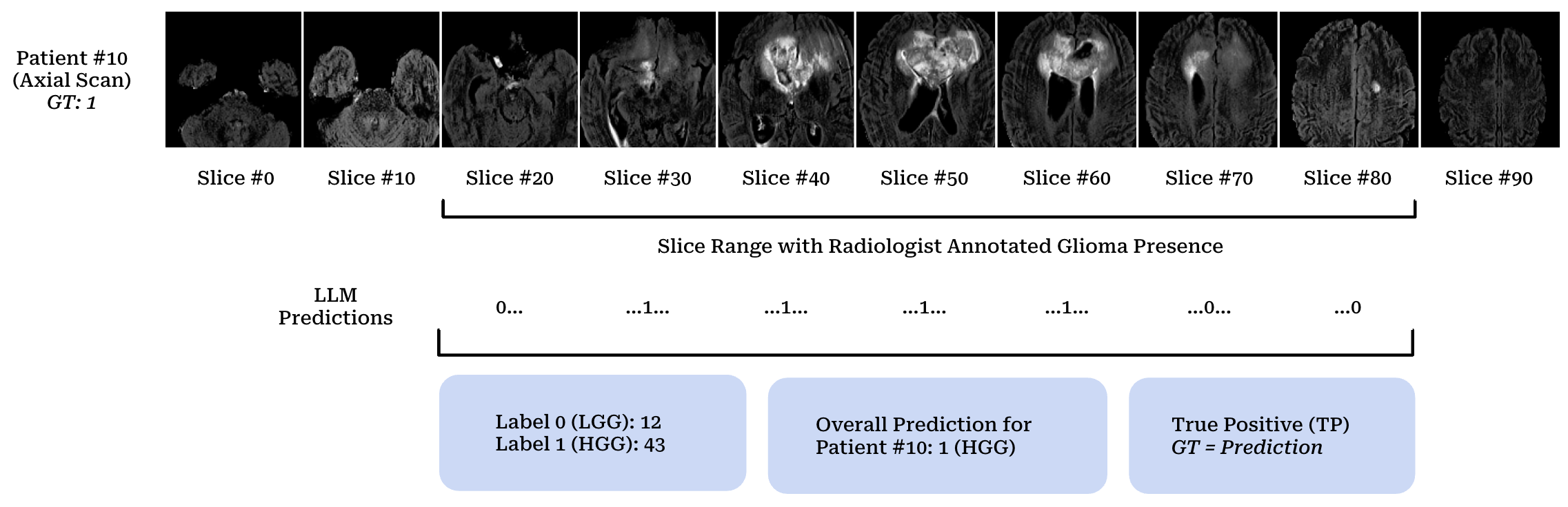}
\caption{Diagram illustrates how predictions for each patient were made by the LLM.}
\label{fig5}
\end{figure}

The LLM approach, by nature, was limited to 2D inputs, as the LLM Vision Model only supports 2D inputs. This constraint, coupled with the slower inference times of the LLM, motivated several key design decisions. Instead of processing the full 3D scan, which would require handling each slice individually, we focused on the axial slices from the FLAIR modality. FLAIR is considered the superior imaging modality for detecting brain anomalies \cite{bakshi_fluid-attenuated_2001}, and limiting the data to this modality helped reduce processing time while still providing high-quality information for the task. Furthermore, because each slice prediction had to be processed separately, we implemented a tallying technique to determine the final patient classification. Specifically, we used a majority vote (winner-takes-all) across all slice predictions for each patient, ensuring that the most consistent label (HGG or LGG) prevailed. This approach is visualized in Figure~\ref{fig5}. For segmentation tasks, we computed evaluation metrics (such as Dice coefficient and Hausdorff distance, discussed later) for each slice independently, before averaging these metrics across the entire patient to obtain a final score. This slice-by-slice evaluation, while a limitation of the LLM approach, was a necessary adaptation given the model’s inability to process the full 3D spatial context. 

\subsubsection{Image classification}
For the image classification, we aimed to classify each patient as either an HGG or LGG patient based on their MRI scans, utilizing all four imaging modalities as a single 3D input (a 95x128x128 image with a depth of four modalities). The corresponding binary ground truth labels were 0 for LGG and 1 for HGG \cite{bakas_identifying_2019, bakas_advancing_2017, menze_multimodal_2015, bakas_dataset, noauthor_brats_lgg}.

\begin{figure}[h!]
\centering
\includegraphics[width=0.6\textwidth]{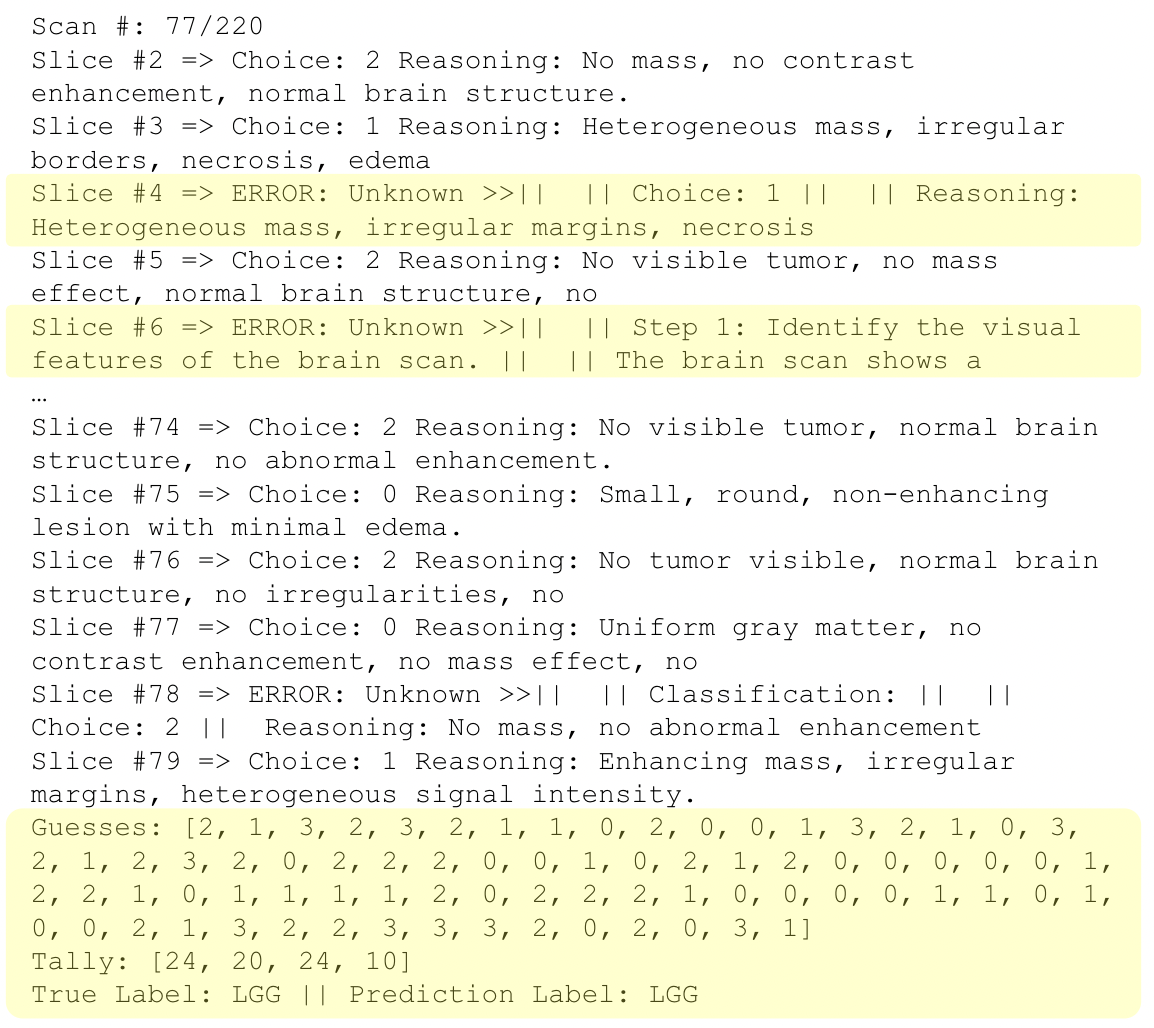}
\caption{Sample response for one patient from the general LLM model. Most responses follow the correct format although there are some outliers. The tally is shown at the bottom and this scan was predicted as LGG.}
\label{fig10}
\end{figure}

\begin{figure}[h!]
\centering
\includegraphics[width=0.8\textwidth]{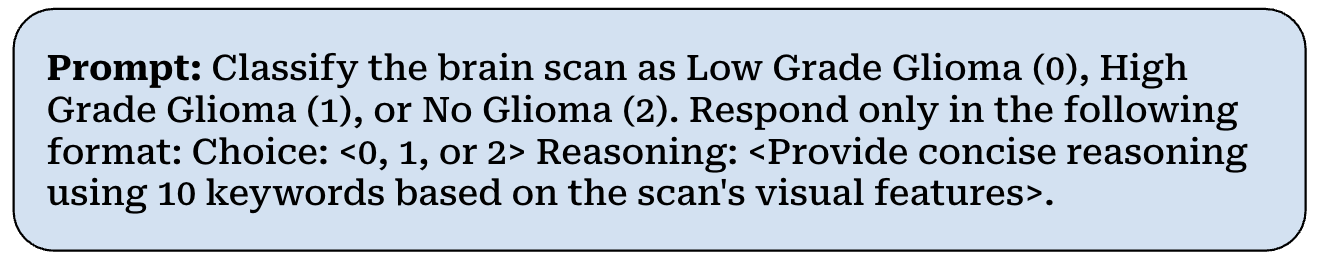}
\caption{LLM general model prompt sample for the classification task.}
\label{fig9}
\end{figure}

The general out-of-the-box LLM, LLaMA 3.2 Instruct, was tested using the same test cohort of 55 patients as the CNN. The segmentation labels were used to select the appropriate axial slice depth range for each patient. Each slice from the FLAIR scans was fed separately to the LLM, which classified the slice as either Low Grade Glioma (LGG-0), High Grade Glioma (HGG-1), or No Glioma (NG-2) based on visual features. The model also responded with concise reasoning using 10 keywords related to the scan’s visual characteristics to ensure the model wasn’t making a random decision. Predictions for each slice were aggregated and the final patient classification was determined using a majority vote (winner-takes-all) across all slice predictions. A sample output response from the general LLM is shown in Figure~\ref{fig10}, and the corresponding prompt is provided in Figure~\ref{fig9} for the classification task.

In addition to the primary objectives comparing the subspecialized LLM, the general LLM, and the baseline CNN in performance, we also conducted a consistency test to evaluate the robustness of the out-of-the-box LLM in this classification task. This involved inputting the same image and prompt into the LLM 95 times, simulating the maximum number of axial slices (95) if a glioma were present throughout all slices, to observe the prediction distribution. A robust model would consistently produce the same prediction for identical inputs. 

\subsubsection{Image segmentation}

For the image segmentation task, we explored three different segmentation approaches to assess the effectiveness of these various segmentation strategies, particularly when transitioning from the CNN baseline models to both the general LLM and the fine-tuned LLM approaches. Similar to the image classification task, we processed the dataset and fine-tuned the model to create a subspecialized version. 

We considered using the Segment Anything Model (SAM) \cite{kirillov_segment_2023}, a general-purpose segmentation model, as the general foundational model. However, we opted not to use SAM for several key reasons. Firstly, SAM relies on explicit user prompts, such as seed points or bounding boxes, which may provide external guidance not typically available to CNN-based models. This explicit prompting introduces additional context that makes direct comparison with CNN-based approaches less valid. Secondly, the ViT architecture used in SAM incorporates self-attention mechanisms, allowing the SAM model to capture long-range dependencies and global context within the image. This feature makes it harder to separate the contributions of spatial image understanding from the impact of prompt-based enhancements in LLMs. Hence, we decided to use LLaMA 3.2 Instruct for the image segmentation task as well. Given that our primary objective is to evaluate the capabilities of a general-purpose vision LLM and compare it to a full CNN baseline model, the use of SAM also would complicate this comparison. Thus, we chose to explore segmentation approaches that provide more efficient and structured representations for the LLM, allowing for better alignment with token limitations while still capturing similar accuracy information, but in a more condensed form compared to a full segmentation mask.

The first segmentation approach was to identify the tumor center point. The goal of this method was to predict the center of the glioma in each MRI scan by identifying the center of the glioma for each axial slice. The ground truth for each slice consisted of the center of glioma (extracted from the bounding box), represented as a single (x, y) coordinate, indicating the x and y coordinates of the glioma's center. This approach simplified the task by focusing on the central location of the glioma, while still providing essential information for glioma localization. The model was expected to predict the center coordinates in the same format for each slice.

The second segmentation approach was to segment the tumor bounding box. The goal of this method was to segment the glioma present in each MRI scan by predicting a bounding box for each axial slice containing the glioma. The ground truth for each slice consisted of a bounding box, represented by the minimum and maximum coordinates along the x and y axes (xmin, xmax, ymin, ymax), forming a square around the glioma. This bounding box was defined by four points: top-left, top-right, bottom-right, and bottom-left, stored in an array of shape (4, 2) for each slice. The model was expected to predict the bounding boxes in the same format for each slice, with the goal of accurately identifying the glioma's location within the image.

The third segmentation approach was to segment a tumor bounding polygon. The goal of this method was to predict the detailed boundaries of the glioma in each MRI scan by representing the glioma as a polygon. For each axial slice containing glioma, the boundary was traced using 10 to 15 points, starting from the north-most point and following a clockwise direction. Smaller gliomas were represented with fewer points, while larger or irregularly shaped gliomas used more. The ground truth for each slice was stored as a list of coordinates with shape (10-15, 2), capturing the precise and tight contours of the glioma. The model was expected to predict the polygon representation of the glioma in the same format, providing a detailed and accurate segmentation of the tumor shape and location.

\section{Experimental setup}

\subsection{Implementation details}

The 3D classification CNN was built with an input shape of (N, 4, 95, 128, 128), where N=16 was the batch size. The first convolutional layer applied 8 filters with a kernel size of 3x3x3, stride of 1, and padding of 1. The second convolutional layer used 16 filters, also with a 3x3x3 kernel, stride of 1, and padding of 1. A max pooling operation with a 2x2x2 kernel and stride of 2 followed each convolutional layer. The fully connected layers included 128 hidden units, with a final output layer for binary classification, distinguishing between LGG and HGG. Rectified Linear Unit activation was used for the hidden layers, and a Sigmoid function was applied to the logits. Learning rates were varied between 1e-4 and 5e-7 to train a suitable and generalising model, and model checkpoints were saved and reloaded during training to ensure robustness. After training and validation, the best model was selected and tested using the separate cohort of 55 patients.

For fine-tuning the LLM for image classification, we used Unsloth due to its simplified user pipeline and optimized features that enable efficient training of LLMs. The brain scan dataset was first balanced to address class imbalances, and each image was formatted into a structured prompt for classification. The pre-trained general vision model was then loaded from Unsloth using the FastVisionModel class, which consisted of vision, language, attention, and MLP layers adapted to the task. LoRA was applied with rank 16 and alpha 16 to update only a small set of parameters, reducing computational overhead while maintaining generalization. Gradient checkpointing was also used to optimize memory usage, allowing for larger batches and more complex architectures without exceeding memory limits. Fine-tuning was conducted using the SFTTrainer class, designed for supervised fine-tuning. A batch size of 16 is used with gradient accumulation over 4 steps to balance memory constraints and model performance. The learning rate was set to 2e-6, selected through experimentation for stable and fast convergence. The AdamW optimizer, with 8-bit precision, accelerated training, which was carried out for a maximum of 100 steps. Upon completion, the model was saved as a LoRA-adapted version to Hugging Face for future testing. The same prompt-based testing procedure used for the general LLM was applied to the specialized model by selecting the fine-tuned version.

The 3D segmentation CNN model was designed for the segmentation of brain tumor images, outputting an upsampled binary mask with a shape of (95, 128, 128). The model took inputs with a shape of (N, 4, 95, 128, 128), where N=24 represented the batch size of patient scans. The first convolutional layer applied 16 filters with a 3x3x3 kernel, a stride, and padding of 1. The second convolutional layer utilized 32 filters with the same kernel size, stride, and padding. Each convolution was followed by a max pooling operation with a 2x2x2 kernel and a stride of 2. The bottleneck consisted of a third convolutional layer with 64 filters. For the decoding component, two transposed convolution layers were used for upsampling, followed by a final convolution layer that output a single-channel segmentation mask. ReLU activation was applied after each convolution and deconvolution, while the output was passed through a sigmoid activation function for binary segmentation. The model was trained for 70 epochs with a learning rate of 0.0001, and model checkpoints were saved regularly to avoid overfitting. The model's performance was evaluated using Dice loss, and after training, the best-performing model was saved for further testing on the same separate cohort of 55 patients used in the classification task. During inference and testing, the predicted masks were thresholded using a tuned parameter of 0.7 to create a binary output before being evaluated.

\begin{figure}[h!]
\centering
\includegraphics[width=0.8\textwidth]{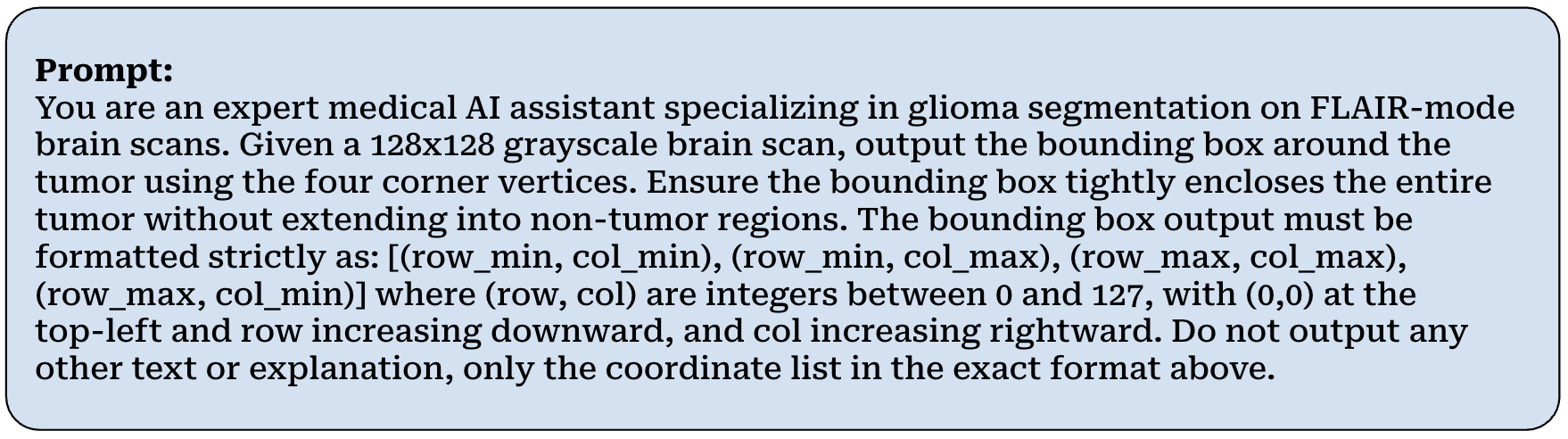}
\caption{LLM model prompt sample for the center point segmentation task. 
}
\label{figa}
\end{figure}

\begin{figure}[h!]
\centering
\includegraphics[width=0.8\textwidth]{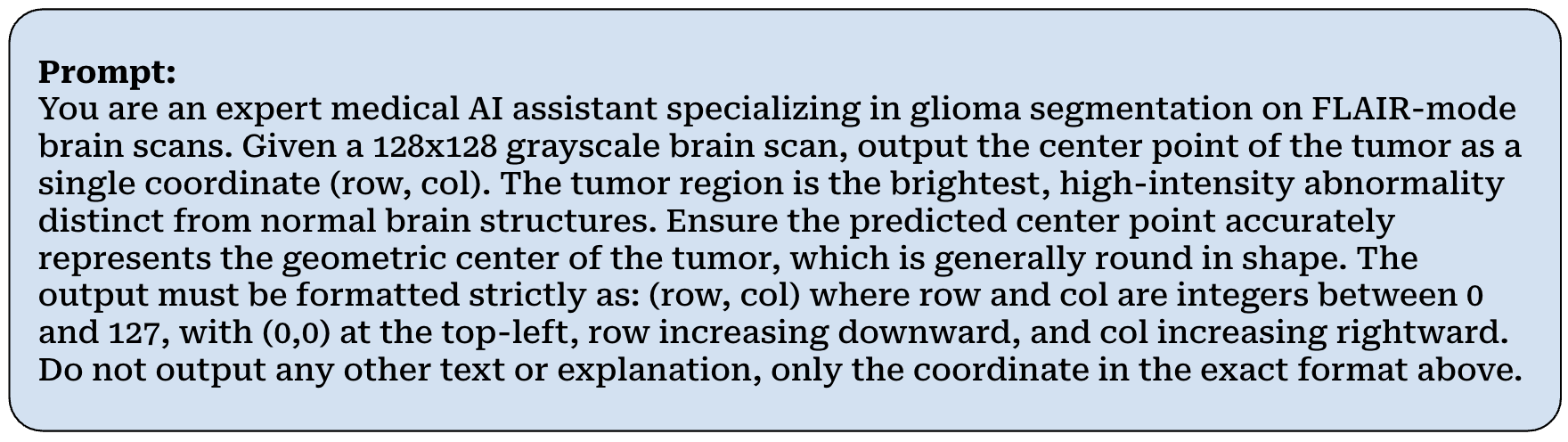}
\caption{LLM model prompt sample for the bounding box segmentation task.
}
\label{figb}
\end{figure}

\begin{figure}[h!]
\centering
\includegraphics[width=0.8\textwidth]{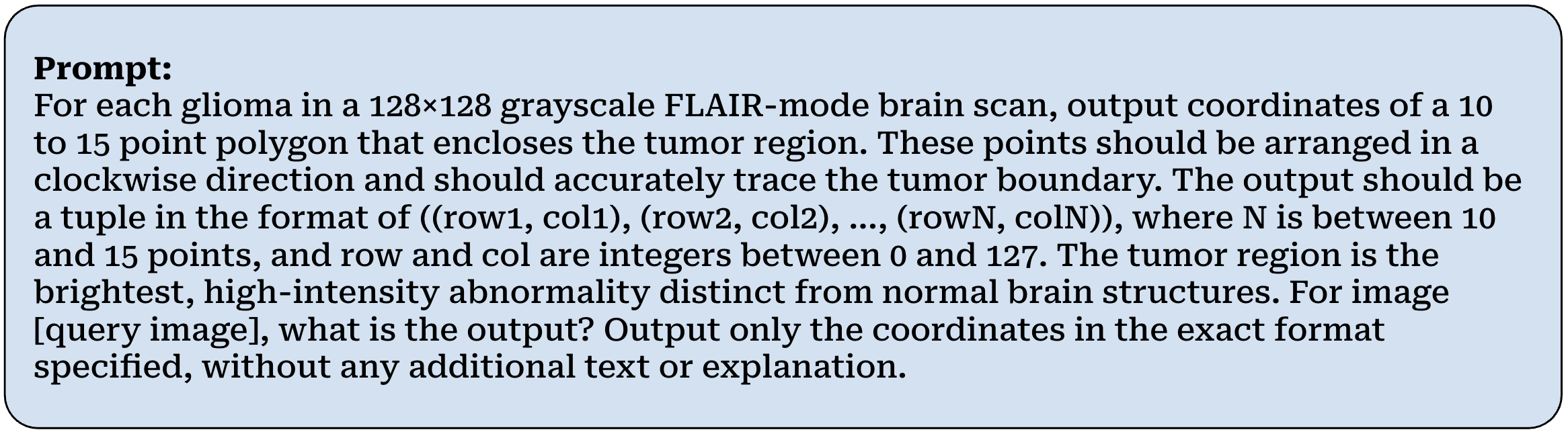}
\caption{LLM model prompt sample for the bounding polygon segmentation task. This prompt was adapted from the methodology used in a previous study that successfully applied prompt-based LLM segmentation. Source: \cite{zhu_llafs_2024}
}
\label{fig_c}
\end{figure}

For image segmentation, the general out-of-the-box LLM was evaluated on the same test cohort of 55 patients used for the CNN baseline. Segmentation labels were utilized to identify the relevant axial slice depth range for each patient. For each case, individual slices from the FLAIR scans were processed sequentially, with the LLM tasked to predict either the tumor’s center point, bounding box (defined by its four corner coordinates), or bounding polygon, depending on the segmentation task. Evaluation metrics were computed independently for each slice and subsequently aggregated across all slices for a given patient to produce a final performance score. While the slice-by-slice evaluation limited the model’s ability to capture the full 3D spatial context, this approach was a necessary compromise due to the LLM’s architectural constraints. The specific prompt used for each segmentation method is illustrated in Figure~\ref{figa}, Figure~\ref{figb}, and Figure~\ref{fig_c}. 

The subspecialized models for image segmentation were fine-tuned using the same general procedure as in the classification task, but adapted for the new segmentation prompts and ground truths. Each image was first formatted into a task-specific prompt, and the dataset was balanced where needed. The LLM was loaded using the Unsloth FastVisionModel, and fine-tuning was applied using LoRA with rank and alpha set to 16, targeting only a subset of parameters to improve efficiency. To support the higher memory demands of the segmentation tasks, gradient checkpointing and gradient accumulation (over four steps) were used, allowing batch sizes between 8 and 24. The learning rate was selected per task to ensure stable convergence, and training was performed for up to 200 steps using the 8-bit AdamW optimizer. After training, the adapted models were saved to Hugging Face and evaluated using the same prompt-based testing strategy as the general LLM.

\subsection{Evaluation metrics}
For the image classification task, accuracy measures the overall correctness of a model by calculating the proportion of correct predictions (true positives and true negatives) out of all predictions made. However, it can be misleading in imbalanced datasets where one class dominates. Precision focuses on the quality of positive predictions by measuring the proportion of true positives among all instances predicted as positive, indicating how many of the positive predictions are actually correct. Recall or sensitivity measures the model’s ability to correctly identify all actual positive cases, highlighting its effectiveness at detecting positive instances. Specificity assesses the model’s capacity to correctly identify negative cases, reflecting the model’s ability to avoid false positives. The F1 score, defined as the harmonic mean of precision and recall, provides a balanced measure, which is especially useful in situations with imbalanced classes. The Area Under the Curve (AUC) evaluates the model’s ability to distinguish between classes across different thresholds, with higher values indicating better performance. AUC cannot be accurately calculated for the LLM, as the model does not generate probabilities. Instead, tally scores were used as a proxy for probability, but this approach lacks credibility and doesn’t offer meaningful insight into model confidence. Together, these metrics offer a comprehensive view of a model’s strengths and weaknesses in classification.

For the segmentation tasks, the primary evaluation metrics used were the Dice coefficient and the 95\% Hausdorff Distance, which provide complementary perspectives on model performance. The Dice coefficient measures the spatial overlap between the predicted segmentation and the ground truth, and is defined as twice the intersection over the sum of the predicted and true volumes. It is particularly useful in medical imaging where high overlap is essential for clinical utility. In contrast, the 95\% Hausdorff Distance captures the spatial agreement at the boundaries by measuring the 95th percentile of distances between the closest points on the predicted and true segmentation surfaces. This metric is more sensitive to boundary errors, making it a valuable complement to Dice, especially in irregular or complex tumor shapes. Additionally, precision, recall, and specificity are computed at the voxel level to provide insight into the model’s ability to correctly identify tumor versus non-tumor regions. While these metrics are also used in classification, their voxel-wise application in segmentation tasks reflects the granularity and spatial precision required in medical image analysis. Together, these measures form a robust evaluation framework for assessing segmentation quality in terms of both overlap and boundary accuracy.

\section{Results}

\subsection{Image classification}

The results of the baseline CNN are presented in Table~\ref{tab:baseline_cnn}. This model was trained with a learning rate of 4e-7 over 200 epochs. After 200 epochs, the model achieved an accuracy of 0.9677, F1 score of 0.9792, precision of 1.0000, recall of 0.9592, and an AUC of 0.9975 on the training set. In the validation set, accuracy dropped to 0.7581, with a slight decrease in F1 score (0.8485) and precision (0.8400), although recall remained strong at 0.8571. The AUC for validation was 0.7221, suggesting moderate classification ability. The testing set yielded an accuracy of 0.8000, F1 score of 0.8706, precision of 0.9024, and recall of 0.8409, with an AUC of 0.8202, showing balanced performance. Specificity was perfect (1.0000) in the training set, but dropped to 0.3846 in validation and increased slightly to 0.6364 in testing, indicating challenges in avoiding false positives in validation and test cohorts.

Several alternative models were also trained but were not selected as the baseline model due to more unstable convergence, as their validation accuracy fluctuated. 

\begin{table}[h]
\centering
\caption{Evaluation of the CNN baseline model (learning rate = 4e-7, epochs = 200).}\label{table1}
\begin{tabular}{cccc}
\toprule
Metric & Training & Validation & Testing \\
\cmidrule(lr){1-1} \cmidrule(lr){2-2} \cmidrule(lr){3-3} \cmidrule(lr){4-4}
Accuracy & 0.9677 & 0.7581 & 0.8000 \\
F1 Score & 0.9792 & 0.8485 & 0.8706 \\
Precision & 1.0000 & 0.8400 & 0.9024 \\
Recall & 0.9592 & 0.8571 & 0.8409 \\
AUC & 0.9975 & 0.7221 & 0.8202 \\
Specificity & 1.0000 & 0.3846 & 0.6364 \\
\bottomrule
\end{tabular}
\label{tab:baseline_cnn}
\end{table}

\begin{figure}[h!]
\centering
\includegraphics[width=\textwidth]{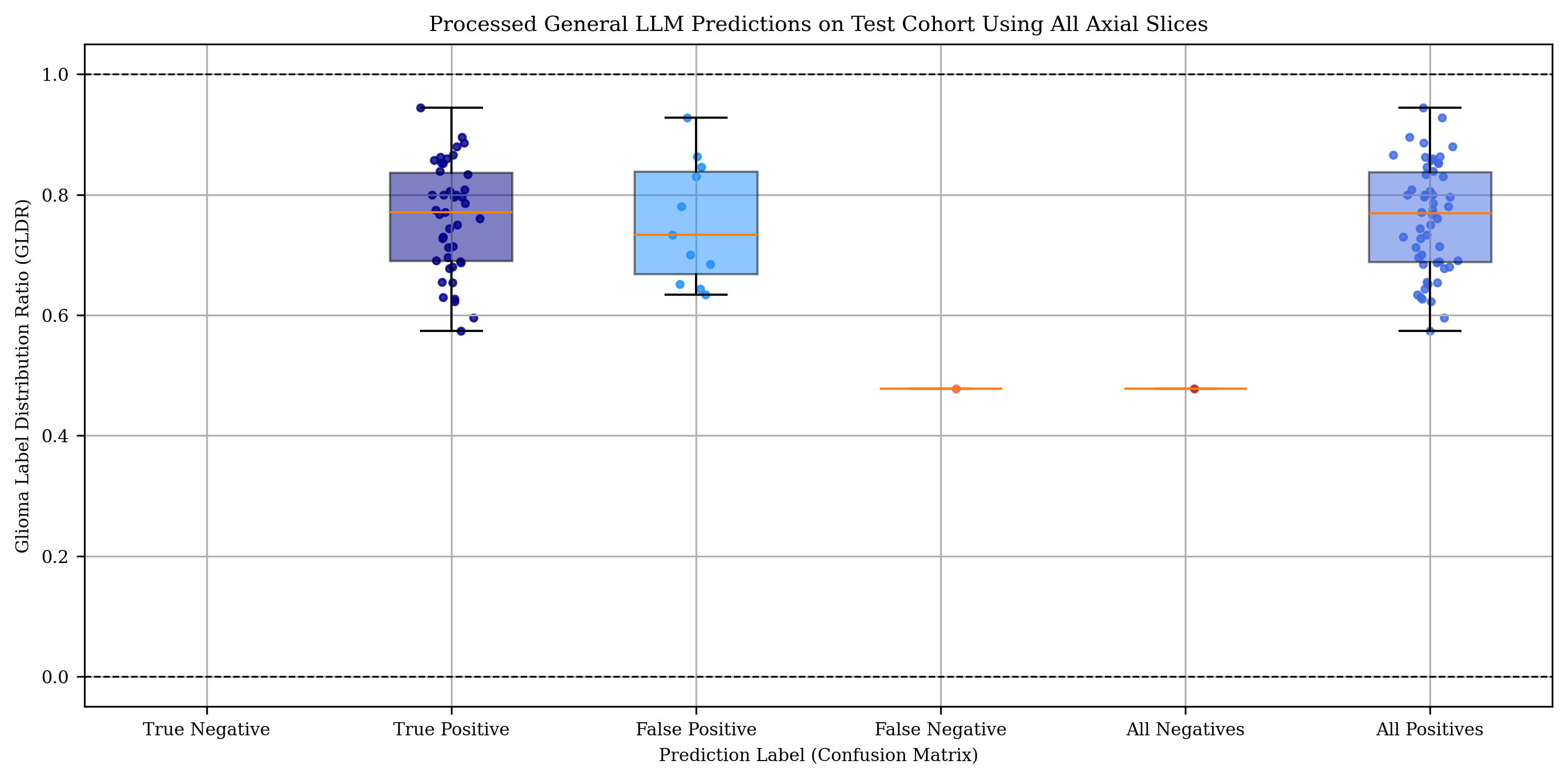}
\caption{Visual representation of the ratio of HGG to LGG labels across each test patient's axial scan slices.
}
\label{fig16}
\end{figure}

\begin{figure}[h!]
\centering
\includegraphics[width=\textwidth]{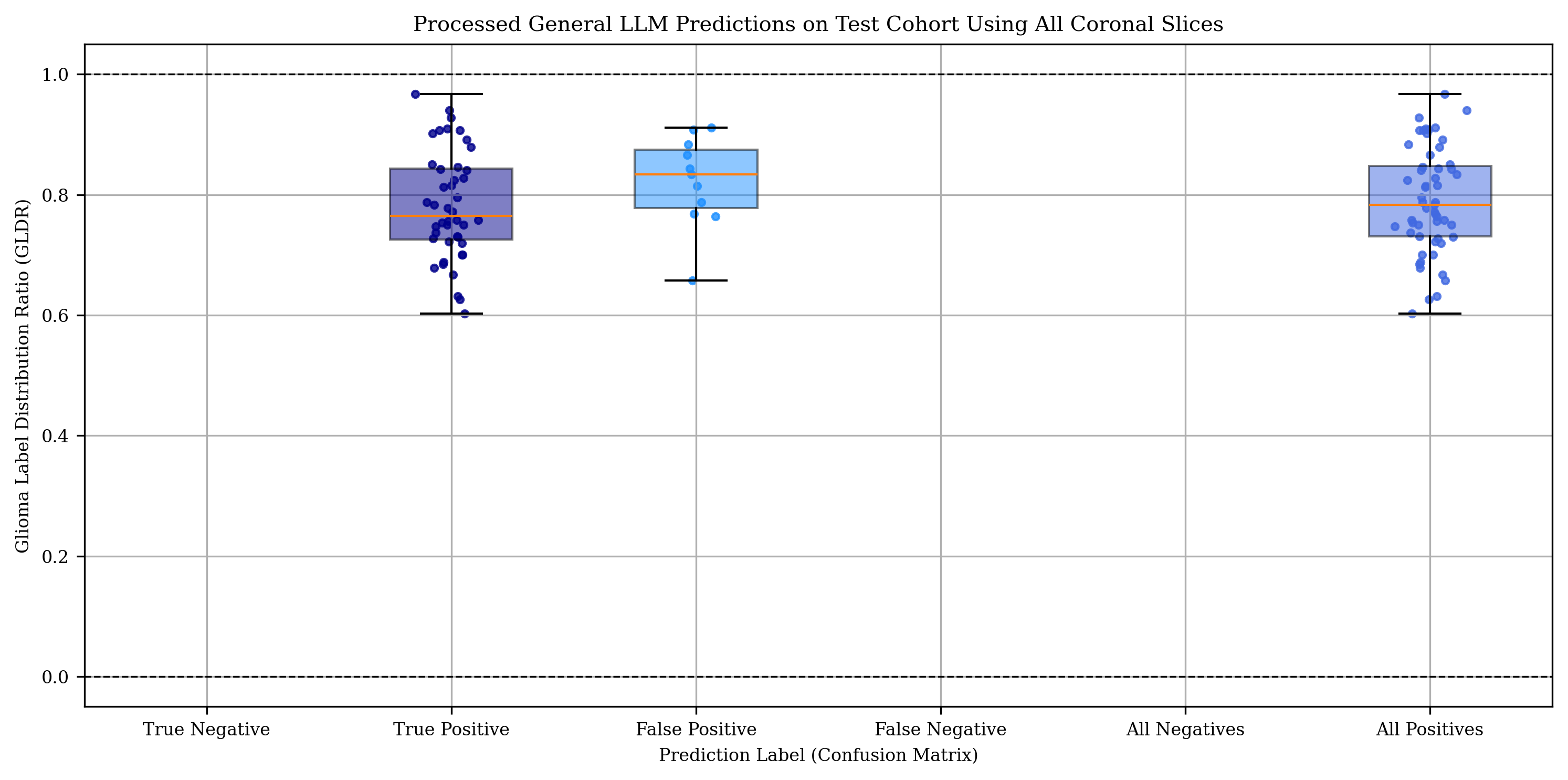}
\caption{Visual representation of the ratio of HGG to LGG labels across each test patient's coronal scan slices.
}
\label{fig17}
\end{figure}

\begin{figure}[h!]
\centering
\includegraphics[width=\textwidth]{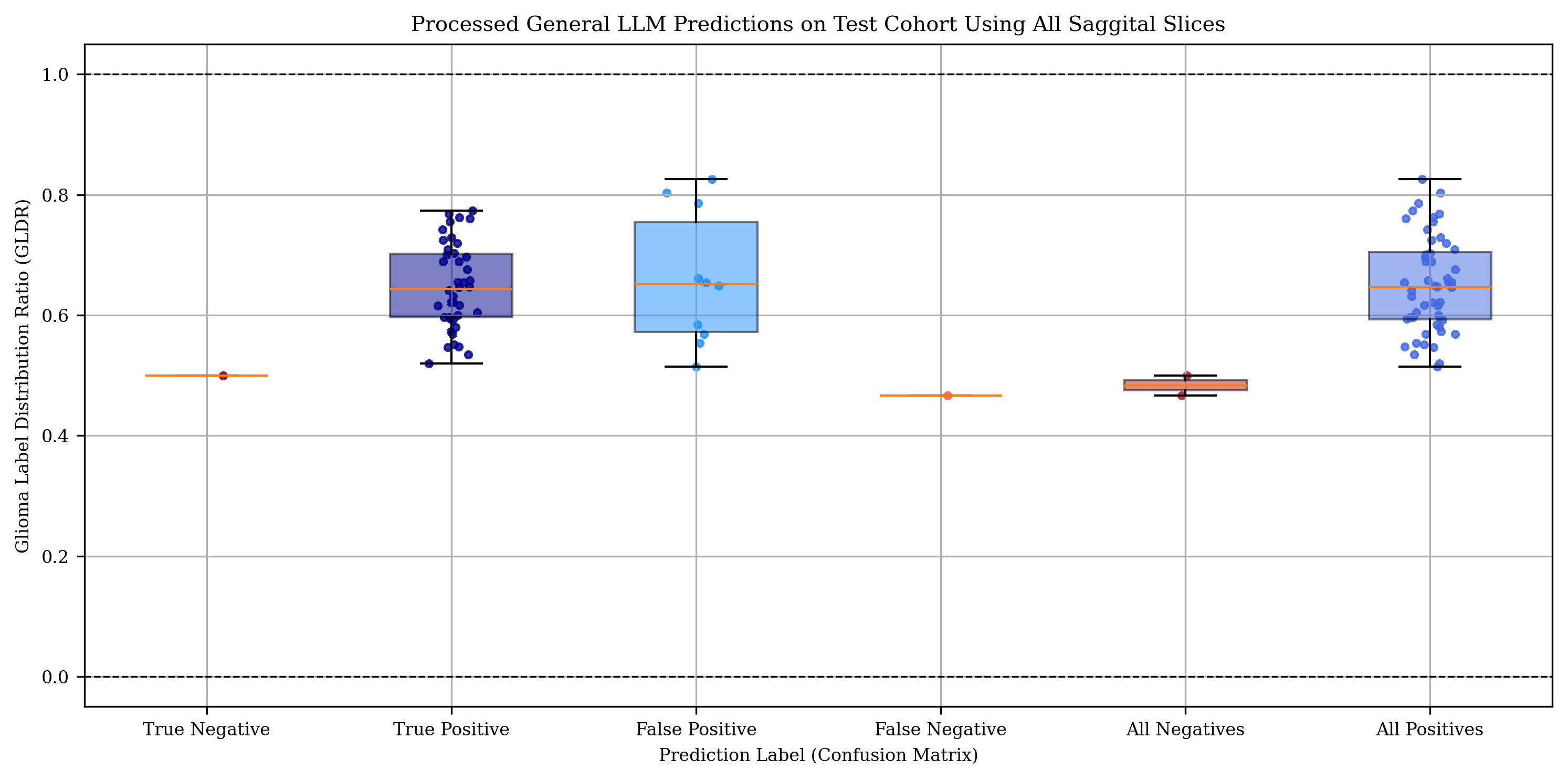}
\caption{Visual representation of the ratio of HGG to LGG labels across each test patient's sagittal scan slices.
}
\label{fig_18}
\end{figure}

The general LLM model's performance was assessed across three different imaging orientations: axial, coronal, and sagittal using the evaluation outlined in the Methodology. The results are presented below in Table~\ref{tab:llm_results}, Figure~\ref{fig16}, Figure~\ref{fig17}, and Figure~\ref{fig_18} respectively. In the figures, the vertical axis is modeled using a Glioma Label Distribution Ratio (GLDR) to highlight the distribution of glioma types across scan slices for each patient for visualization. For a scan with 100 slices, if 25 slices are labeled as LGG and 75 as HGG, the patient would have a ratio of 0.75, indicating the proportion of HGG-labeled slices compared to the total number of slices. 

\begin{table}[h]
\centering
\caption{Evaluation of the general LLM model across axial, coronal, and sagittal imaging orientations during testing.}
\begin{tabular}{cccc}
\toprule
Metric & Axial & Coronal & Sagittal \\
\cmidrule(lr){1-1} \cmidrule(lr){2-2} \cmidrule(lr){3-3} \cmidrule(lr){4-4}
Accuracy & 0.7818 & 0.8000 & 0.7963 \\
F1 Score & 0.8776 & 0.8889 & 0.8842 \\
Precision & 0.7963 & 0.8000 & 0.8077 \\
Recall & 0.9773 & 1.0000 & 0.9767 \\
Specificity & 0.0000 & 0.0000 & 0.0909 \\
\bottomrule
\end{tabular}
\label{tab:llm_results}
\end{table}

The model demonstrated strong recall and accuracy across all orientations, with the coronal achieving the highest accuracy (0.8000) and perfect recall (1.0000), while the axial and sagittal orientations also exhibited strong recall values (0.9773 and 0.9767, respectively). At first glance, these metrics suggest good performance. However, the performance is misleading due to the class imbalance in the dataset, where the majority of slices are HGG, and far fewer are LGG. When examining the data in the figures, it becomes evident that the model predominantly predicts HGG across all cases. This is in line with the imbalance, as the model appears to simply and consistently predict the majority class (HGG) rather than differentiating between HGG and LGG as shown by the zero specificity across all slice orientations. Certain LGG scans demonstrated ratios in the 90\% range, indicating failure of the general LLM model to differentiate between the classes.

Overall, the results indicate that while the model can be said to be accurate in detecting HGG, it is not effectively identifying LGG patients, and is heavily biased in favour of the imbalance in the dataset, hence the high accuracy, precision, recall, and F1 score. These results reflect the performance of the general LLM model. 

\begin{figure}[h!]
\centering
\includegraphics[width=\textwidth]{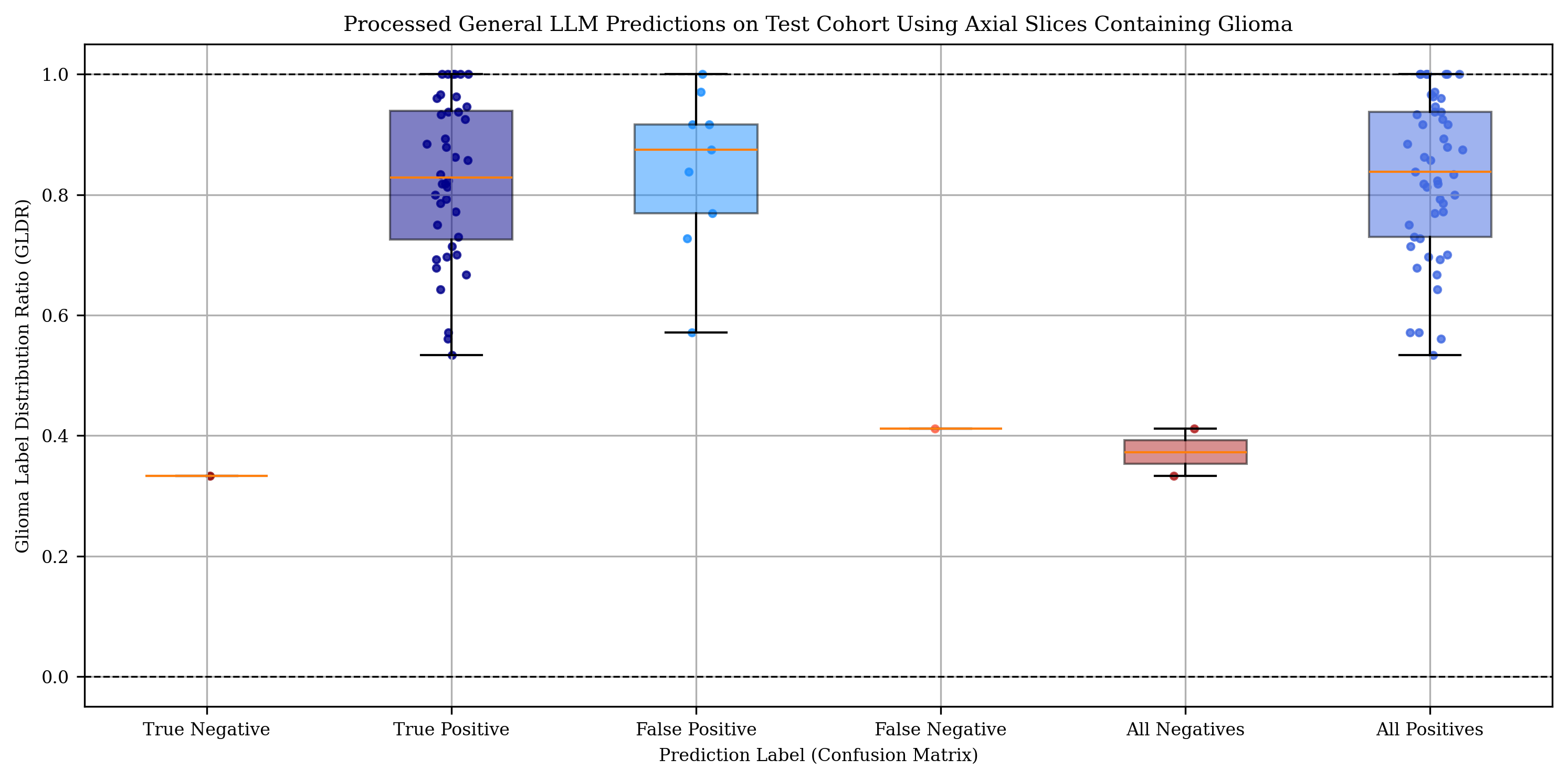}
\caption{Visual representation of the ratio of HGG to LGG labels across test patient axial scan slices with glioma evaluated using the small general LLM model.
}
\label{fig18_smallgen}
\end{figure}

\begin{figure}[h!]
\centering
\includegraphics[width=\textwidth]{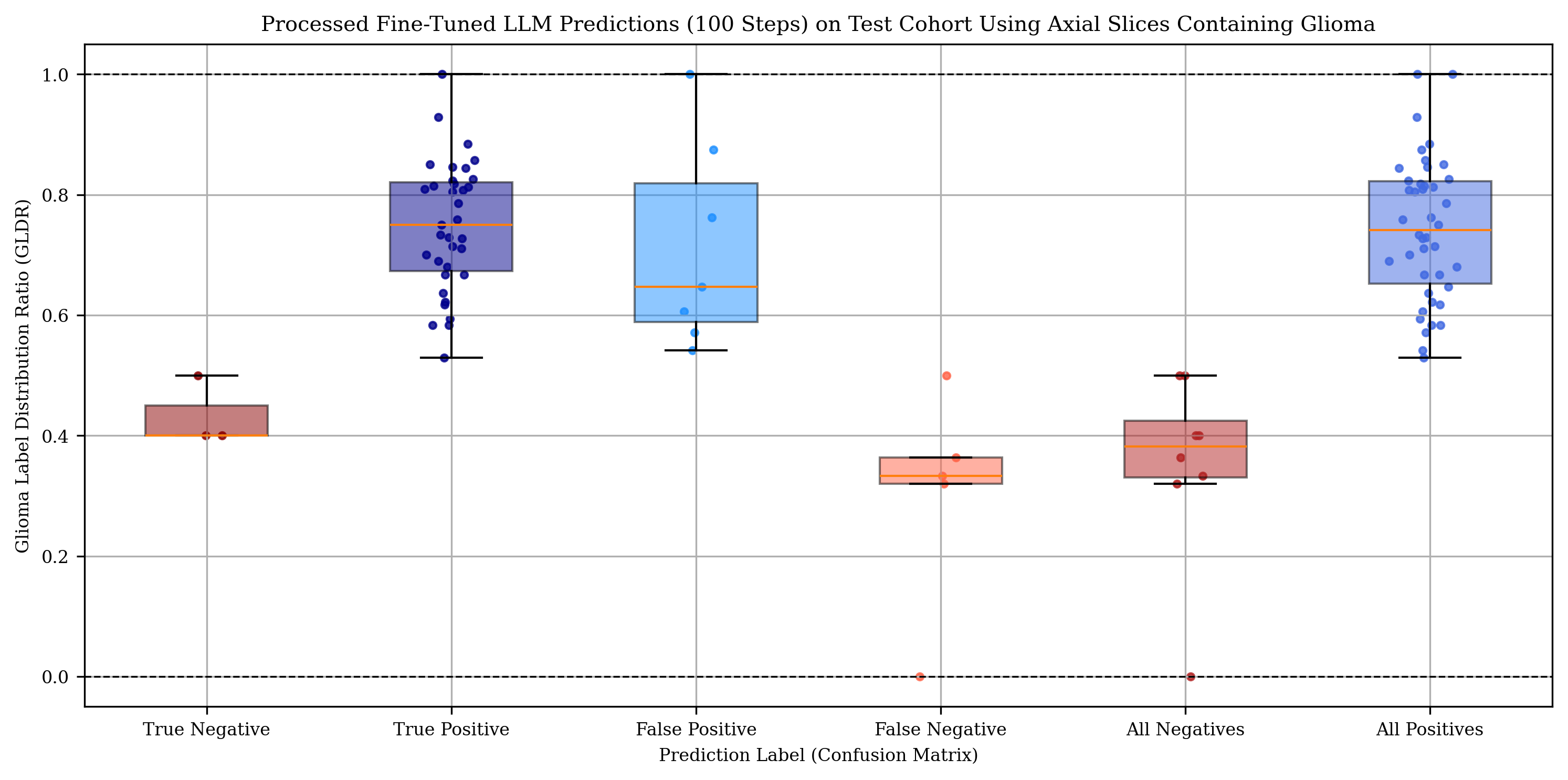}
\caption{Visual representation of the ratio of HGG to LGG labels across test patient axial scan slices with glioma evaluated using the fine-tuned LLM model of 100 steps.
}
\label{fig18_x100}
\end{figure}

\begin{figure}[h!]
\centering
\includegraphics[width=\textwidth]{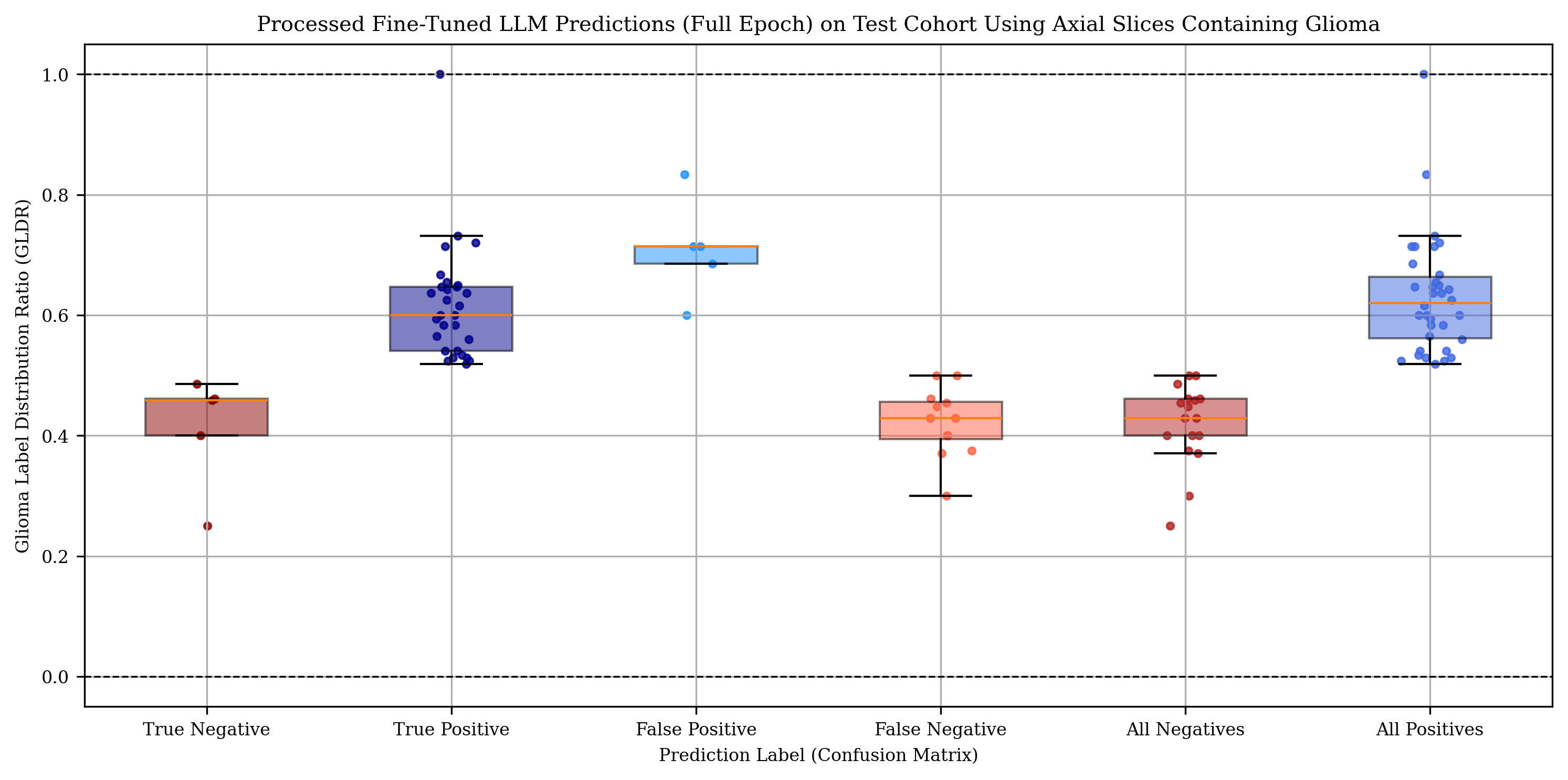}
\caption{Visual representation of the ratio of HGG to LGG labels across test patient axial scan slices with glioma evaluated using the fine-tuned LLM model of one epoch.
}
\label{fig18_xfull}
\end{figure}

The LLM was fine-tuned under two separate conditions. The first was fine-tuning for 100 steps with a batch size of 4, the second was fine-tuning for one full epoch using a batch size of 16. The fine-tuned LLM model was evaluated solely on the axial imaging orientation, using the same methodology as the general LLM to maintain consistency while reducing computational demands. Because the fine-tuning process was performed exclusively on slices containing glioma, the fine-tuned models were also evaluated only on this subset. To ensure a fair comparison, the general LLM was assessed on the same glioma containing slices (referred to as the "General LLM (small)") to ensure that performance is compared on equivalent data across models. The results are once again, visualized using the Glioma Label Distribution Ratio (GLDR), which represents the proportion of glioma types averaged across slices per patient. These are shown in Figure~\ref{fig18_smallgen} for the small general LLM, Figure~\ref{fig18_x100} for the fine-tuned model trained with 100 steps, and Figure~\ref{fig18_xfull} for the model fine-tuned for one full epoch.

\begin{table}[h]
\centering
\caption{Evaluation of the fine-tuned LLM models in comparison to the general LLM.}\label{table5}
\begin{tabular}{cccccc}
\toprule
Model & Accuracy & F1 Score & Precision & Recall & Specificity \\
\cmidrule(lr){1-1} \cmidrule(lr){2-2} \cmidrule(lr){3-3} \cmidrule(lr){4-4} \cmidrule(lr){5-5} \cmidrule(lr){6-6}
General LLM (small) & 0.7636 & 0.8602 & 0.8163 & 0.9091 & 0.1818 \\
Fine-tuned LLM (100 Steps) & 0.76 & 0.8537 & 0.8333 & 0.875 & 0.3 \\
Fine-tuned LLM (Full Epoch) & 0.6667 & 0.7733 & 0.8529 & 0.7073 & 0.5 \\
\bottomrule
\end{tabular}
\end{table}

The results for the full-epoch and 100-step fine-tuned models show some mixed outcomes. For the full-epoch model, there was a noticeable decline in performance, with an accuracy of 0.6667 and an F1 score of 0.7733. The recall dropped to 0.7073, indicating a reduction in the model's ability to identify true positives compared to the 100-step model, which achieved a higher recall of 0.875. The specificity for the full-epoch model was 0.5, suggesting that it still struggled with distinguishing between HGG and LGG despite the fine-tuning effort. On the other hand, the 100-step fine-tuned model showed relatively stable performance, with accuracy at 0.76 and an F1 score of 0.8537, but the precision remained low at 0.8333. Both models demonstrated only marginal improvements over the general model. The full-epoch model, in particular, did not perform as expected, as it failed to improve its ability to differentiate between the two glioma types. The LGG predictions were expected to cluster closer to a ratio of 0-0.1 and the HGG predictions to be near 0.9-1, but this was not the case. Instead, the models produced predictions that were roughly split between HGG and LGG, suggesting that neither model has learned a clear distinction between the two classes and seems to treat the two types as interchangeable, rather than accurately distinguishing between them. While both fine-tuned models showed some marginal improvement over the general LLM, the results were not as promising as anticipated.

For the consistency test, the general LLM model consistently predicted the HGG image as HGG across all 95 trials with the same prompt, which demonstrates good consistency. However, 77 of the 95 trials with the LGG image were incorrectly predicted as HGG, while only 18 were correctly identified as LGG. This led to an overall prediction of HGG, which is incorrect. The split in predictions, especially with the same input, raises concerns about the model's consistency, as it flipped between predictions despite identical input. This behavior is unusual and indicates a lack of stability in the general model's response. The fine-tuned LLM results show that for the HGG image, the model incorrectly predicted 23 out of 95 trials as LGG and 72 as HGG, which is worse compared to the general LLM's performance. For the LGG image, 42 trials were predicted as LGG and 53 as HGG, resulting in a near 50-50 split. This outcome is somewhat expected because as the results above show, the model has not learned to distinguish between LGG and HGG effectively. The inconsistent predictions, especially for the LGG case, suggest that fine-tuning did not enhance the model's ability to differentiate the classes and there is still a need for alternative and improved fine-tuning strategies.

\subsection{Image segmentation}

\begin{figure}[h!]
\centering
\includegraphics[width=\textwidth]{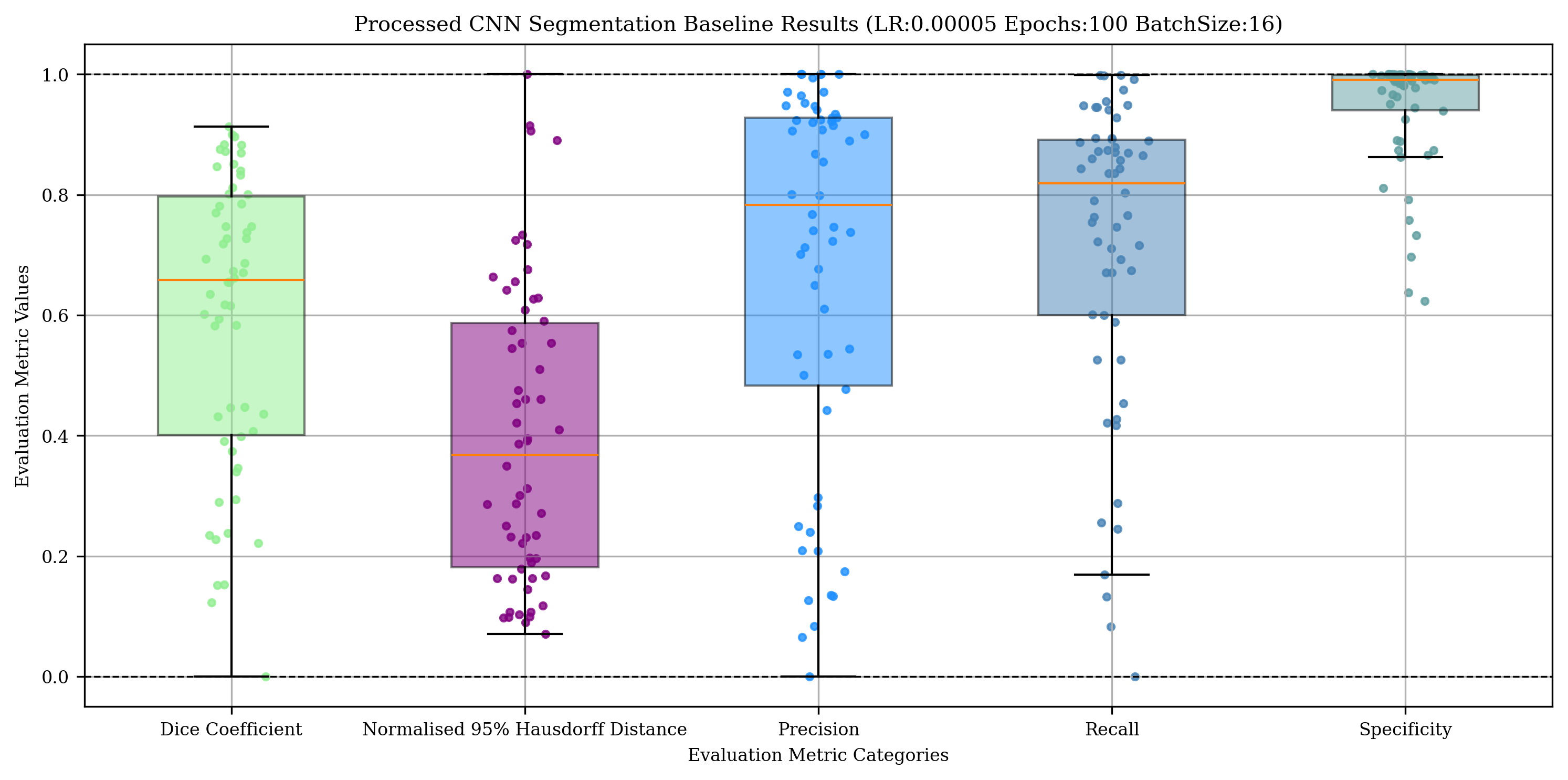}
\caption{Evaluation metrics for the CNN baseline model predictions.
}
\label{fig_28}
\end{figure}

The CNN segmentation baseline was trained with various configurations, ultimately achieving the best results with 100 epochs and a batch size of 16. Table~\ref{tab:baseline_cnn_seg} and Figure~\ref{fig_28} visualizes the evaluation metrics across test patients.

\begin{table}[h]
\centering
\caption{Evaluation of the CNN baseline model (learning rate = 5e-5, epochs = 100).}
\label{table2}
\begin{tabular}{lccc}
\toprule
Metric & Training & Validation & Testing \\
\cmidrule(lr){1-1} \cmidrule(lr){2-2} \cmidrule(lr){3-3} \cmidrule(lr){4-4}
Dice Coefficient & 0.6020 & 0.6158 & 0.5942 \\
95\% Hausdorff Distance & 53.9291 & 57.9943 & 50.5580 \\
Precision & 0.7011 & 0.7225 & 0.6765 \\
Recall & 0.7220 & 0.7109 & 0.7163 \\
Specificity & 0.9251 & 0.9506 & 0.9445 \\
\bottomrule
\end{tabular}
\label{tab:baseline_cnn_seg}
\end{table}

After 100 epochs, the model achieved a Dice coefficient of 0.6020, a 95\% Hausdorff Distance of 53.9291, precision of 0.7011, recall of 0.7220, and specificity of 0.9251 on the training set. The validation set yielded a Dice coefficient of 0.6158, a 95\% Hausdorff Distance of 57.9943, precision of 0.7225, recall of 0.7109, and specificity of 0.9506. On the testing set, the model achieved a Dice coefficient of 0.5942, a 95\% Hausdorff Distance of 50.5580, precision of 0.6765, recall of 0.7163, and specificity of 0.9445. These results show a consistent performance across training, validation, and testing sets, with the model demonstrating reliable segmentation accuracy, good tumor boundary alignment, and effective detection of tumor regions while maintaining a strong ability to identify non-tumor areas. Other baseline CNN models were also trained but not selected due to suboptimal convergence, indicated by slight plateaus during training, and had Dice testing scores ranging from 55\% to 57\%. 
 
To qualitatively assess the performance of the CNN baseline model, we visualized its segmentations on the 55 patients from the testing cohort. The results, displayed in the figures below, highlight both successes and challenges. In each figure, the ground truth segmentation is shown in blue, while the CNN model's predictions are represented in red. This visualization offers a clear overview of the model's performance, showcasing areas where it performs well and identifying regions where it struggles.

\begin{figure}[h!]
\centering
\includegraphics[width=\textwidth]{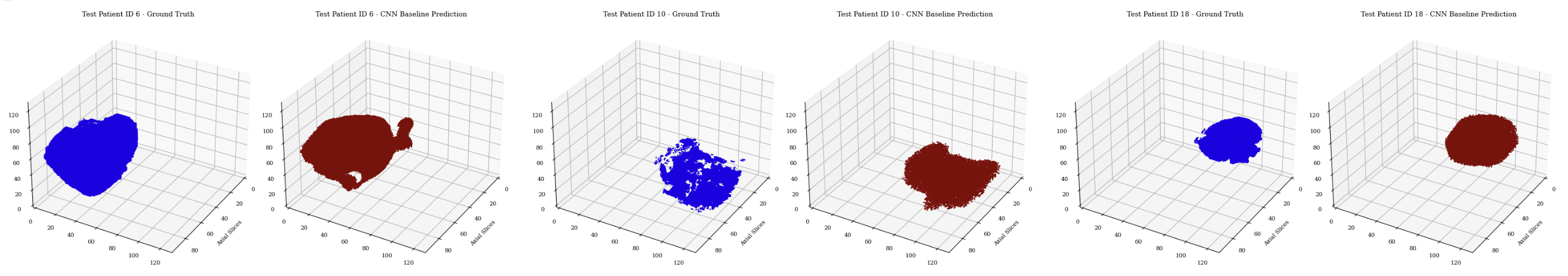}
\caption{
Baseline CNN segmentation visualization showed accurate glioma segmentation with correct size, location, and voxel boundaries.
}
\label{fig_base1}
\end{figure}

\begin{figure}[h!]
\centering
\includegraphics[width=\textwidth]{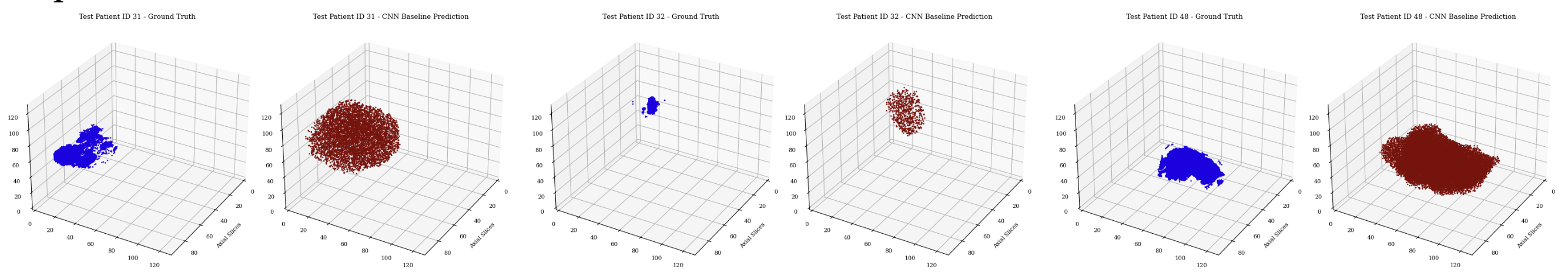}
\caption{Baseline CNN segmentation visualization showed accurate glioma segmentation, but with less precise tumor boundaries. 
}
\label{fig_base2}
\end{figure}

\begin{figure}[h!]
\centering
\includegraphics[width=\textwidth]{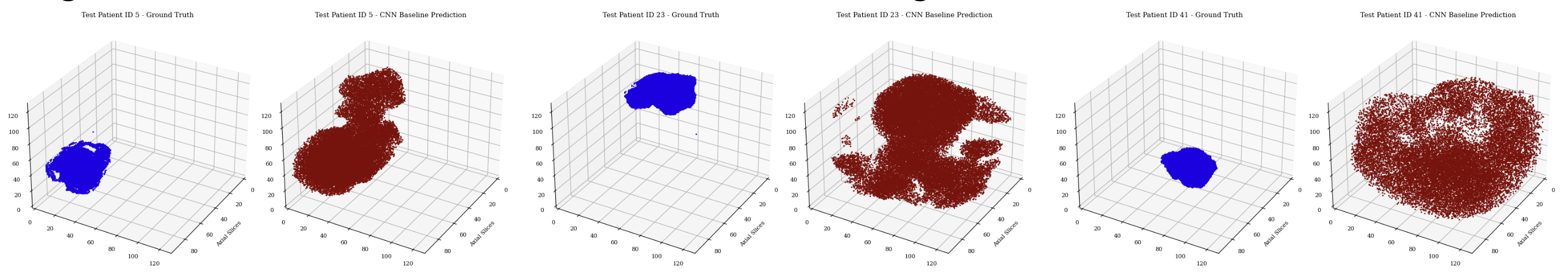}
\caption{Baseline CNN segmentation visualization showed overestimation of glioma size, capturing broader brain regions or the entire brain.
}
\label{fig_base3}
\end{figure}

Figure~\ref{fig_base1} illustrates that the model accurately segmented gliomas, with segmentations closely aligning with the ground truth. The model effectively captured the tumor’s size, location, and voxel boundaries, reflecting strong performance in these instances by maintaining the general shape and spatial distribution of the glioma. In contrast, Figure~\ref{fig_base2} demonstrates that while the segmentations for other gliomas were still largely accurate, some limitations emerged. Although the model successfully identified the correct regions and preserved the general 3D shape of the brain, it struggled with the precision of tumor boundaries. In these cases, the predicted pixels tended to remain around the correct regions, suggesting that while the overall segmentation was generally accurate, finer tumor details were missed. There were also instances where the model overestimated the glioma size, as seen in Figure~\ref{fig_base3}, where it captured broader regions of the brain, sometimes even encompassing the entire brain. This over-segmentation could be attributed to poorer contrast in the images, which made the glioma less distinct, or the presence of other brain anomalies that the model mistakenly identified as part of the glioma. In such cases, the segmentation was less focused on the tumor itself and more on surrounding brain structures, resulting in inaccurate tumor boundaries.

\begin{figure}[h!]
\centering
\includegraphics[width=\textwidth]{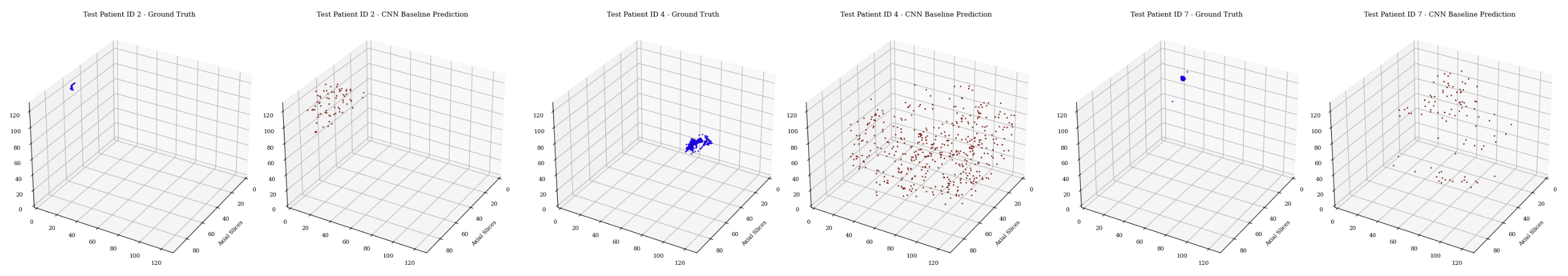}
\caption{Baseline CNN segmentation visualization showed challenges in detecting smaller, irregularly shaped gliomas.
}
\label{fig_base4}
\end{figure}

\begin{figure}[h!]
\centering
\includegraphics[width=\textwidth]{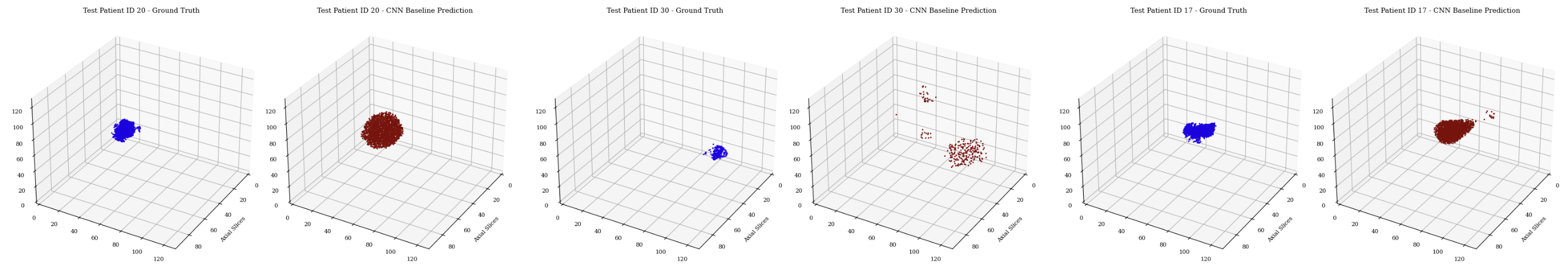}
\caption{Baseline CNN segmentation visualization showed accurate detection of smaller gliomas in potentially higher-quality scans.
}
\label{fig_base5}
\end{figure}

Smaller gliomas, particularly those with irregular shapes, presented a significant challenge for the model, as shown in Figure~\ref{fig_base4}. In many cases, the model struggled to detect these tumors, often predicting a scattered cloud of points rather than a well-defined glioma. This issue may have stemmed from the model either over-segmenting the entire brain or failing to distinguish small gliomas from surrounding tissues. As a result, these instances resulted in lower accuracy, with the model having difficulty precisely localizing the tumor. However, some smaller gliomas were accurately detected, as seen in Figure~\ref{fig_base5}, indicating that the model's performance was potentially influenced by the quality of the MRI scan.

\begin{figure}[h!]
\centering
\includegraphics[width=\textwidth]{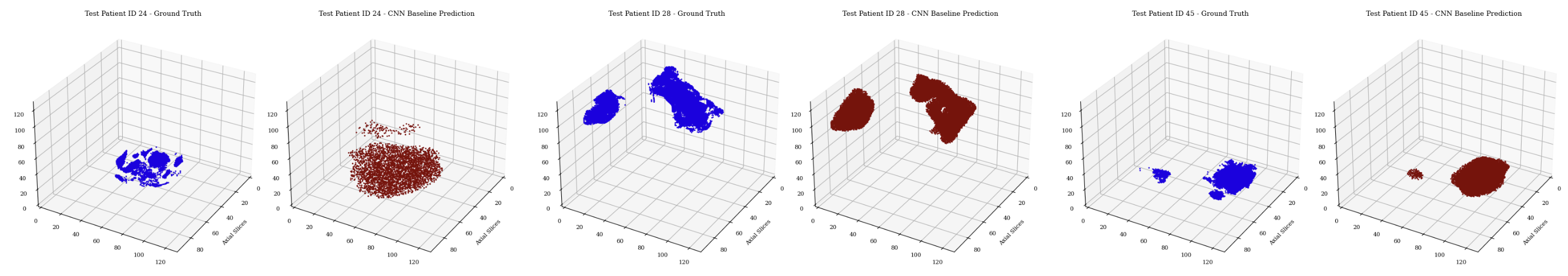}
\caption{Baseline CNN segmentation visualization showed accurate identification and segmentation of complex gliomas with unusual or multiple boundaries.
}
\label{fig_base6}
\end{figure}

\begin{figure}[h!]
\centering
\includegraphics[width=\textwidth]{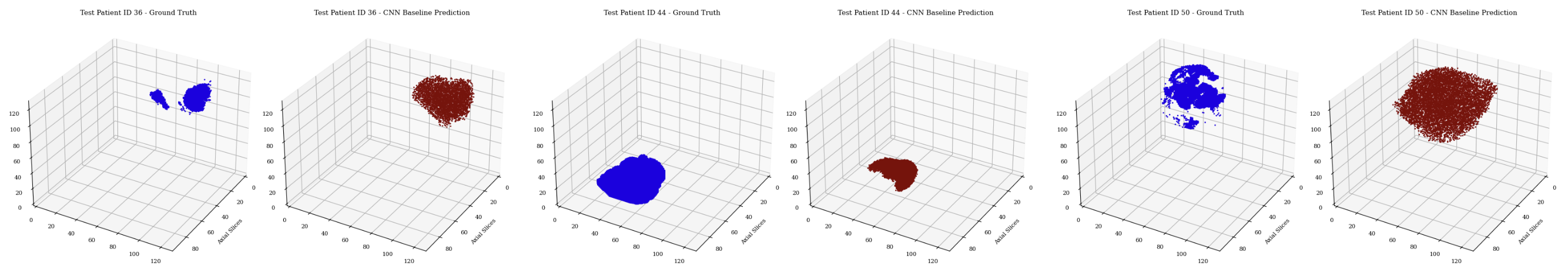}
\caption{Baseline CNN segmentation visualization showed incomplete segmentation of complex gliomas with irregular shapes.
}
\label{fig_base8}
\end{figure}

For cases where the glioma had unusual boundaries or where multiple gliomas were present, as shown in Figure~\ref{fig_base6}, the CNN baseline model still demonstrated strong performance in accurately identifying and segmenting these complex scenarios. The model effectively predicted the boundaries of each glioma, even in the presence of multiple or irregularly shaped tumors. However, in other cases, such as in Figure~\ref{fig_base8}, the model struggled to capture the full extent of complex gliomas, especially those with irregular shapes. These segmentations often resulted in smaller areas being captured than the actual size of the glioma, indicating that the model still faced difficulties with accurately delineating complex tumor boundaries.

For the center point segmentation task, the performance of the general LLM model was evaluated using only the axial imaging orientation. We also used different evaluation metrics as substitutes for the Dice coefficient and 95\% Hausdorff distance. Instead of the Dice coefficient, we calculated the percentage of predicted points within the ground truth bounding box for each slice, as an area-related proxy. For the 95\% Hausdorff distance, we used the Euclidean distance between the predicted and ground truth center points and took the 95th percentile of all Euclidean distances for each patient. We also measured the 95th percentile of the closest distances from the predicted center point to the ground truth bounding box, providing a metric that accounts for tumor size. All distance-based metrics were normalized to the 128x128 image size. The results are presented below in Figure~\ref{fig_segcen_gen}.

\begin{figure}[h!]
\centering
\includegraphics[width=\textwidth]{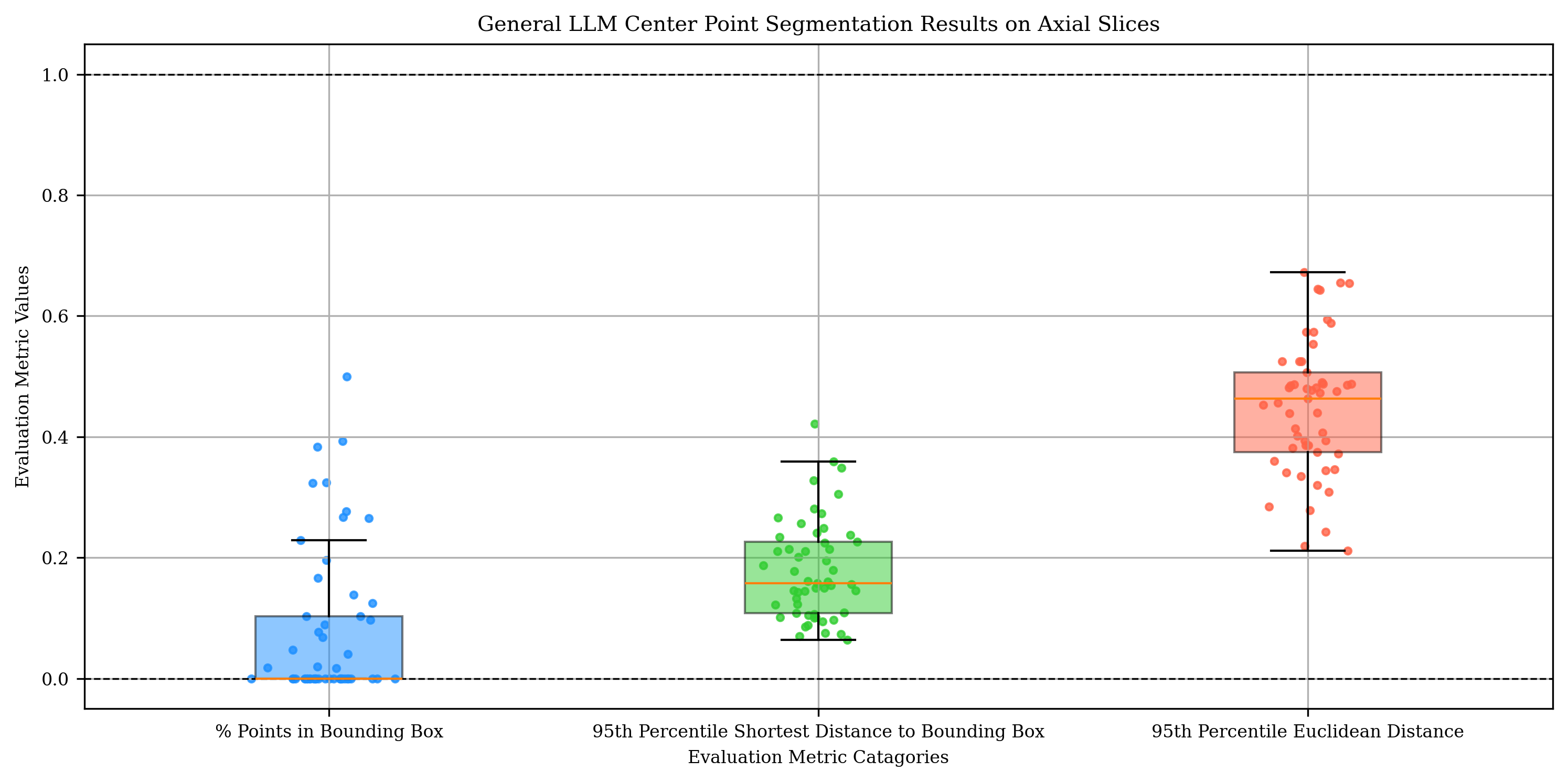}
\caption{Evaluation metrics for the general LLM center point model predictions.
}
\label{fig_segcen_gen}
\end{figure}

\begin{table}[h!]
\centering
\caption{Evaluation of the general LLM center point model.}
\label{table3}
\begin{tabular}{lc}
\toprule
Metric & Value \\
\cmidrule(lr){1-1} \cmidrule(lr){2-2}
Average \% Points in Bounding Box & 0.0805 \\
Average 95\textsuperscript{th} Percentile Shortest Distance to Bounding Box & 22.8566 \\
Average 95\textsuperscript{th} Percentile Euclidean Distance & 57.4112 \\
\bottomrule
\end{tabular}
\end{table}

\begin{figure}[h!]
\centering
\includegraphics[width=\textwidth]{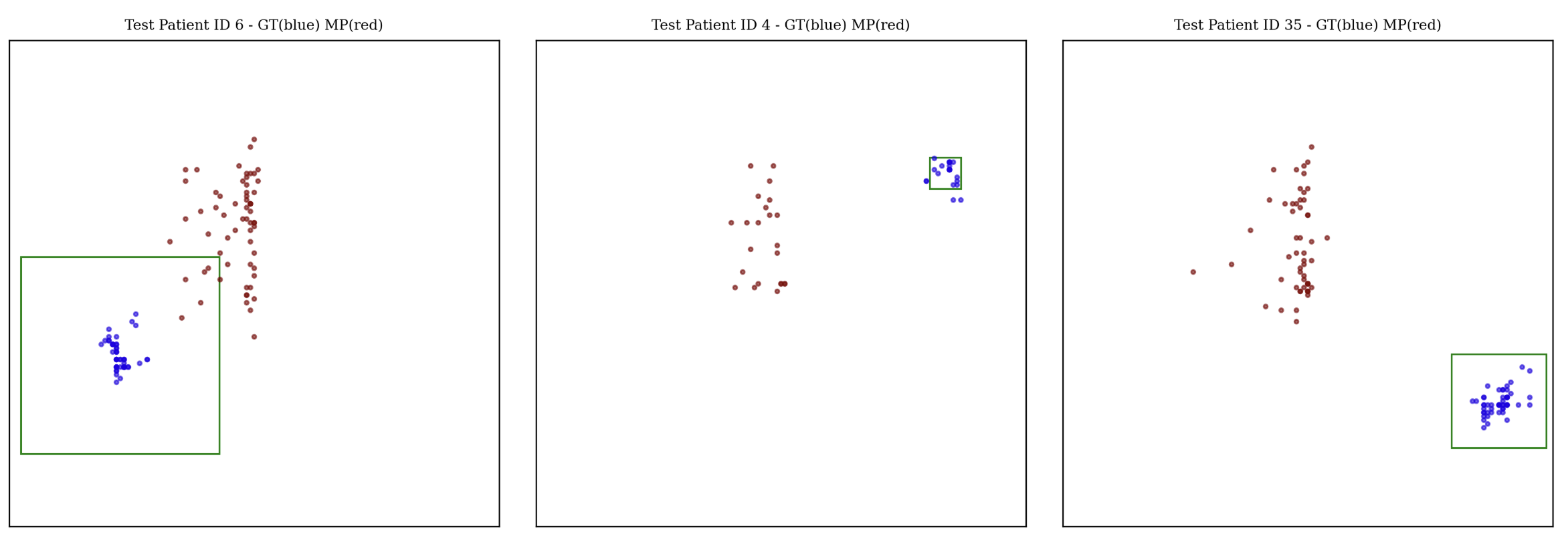}
\caption{General LLM center point segmentation visualization showed, for all patients, the model predicted points near the center of the frame with some variance, regardless of the ground truth glioma location.
}
\label{fig_aaa}
\end{figure}

The results of the general LLM model on center point segmentation were not promising. Ideally, a high percentage of predicted points should fall within the ground truth bounding box, but the median value was close to zero, indicating that most predicted points were not aligned with the true tumor location. This suggests that the accuracy of the center point segmentation was largely random, as seen in Figure~\ref{fig_aaa}. The model tended to predict points near the center of the frame with some variance, contributing to the low percentage of points within the bounding box. For smaller or more distant gliomas, predicted points were often outside the box, resulting in a median percentage near zero. In terms of distance metrics, the model’s predictions were, on average, about half the frame's distance away from the true tumor location. The 95th percentile shortest distance to the bounding box averaged 22.8566, while the 95th percentile Euclidean distance was 57.4112. These distances reflect the model's tendency to predict points near the image center, rather than accurately localizing tumor centers. This pattern suggests that the model is not effectively differentiating between different gliomas, regardless of their size or location, which is again reflected in the visualizations in Figure~\ref{fig_aaa}. The predictions are clustered near the center, as the model is failing to identify relevant features within the image to guide tumor localization. 

\begin{figure}[h!]
\centering
\includegraphics[width=\textwidth]{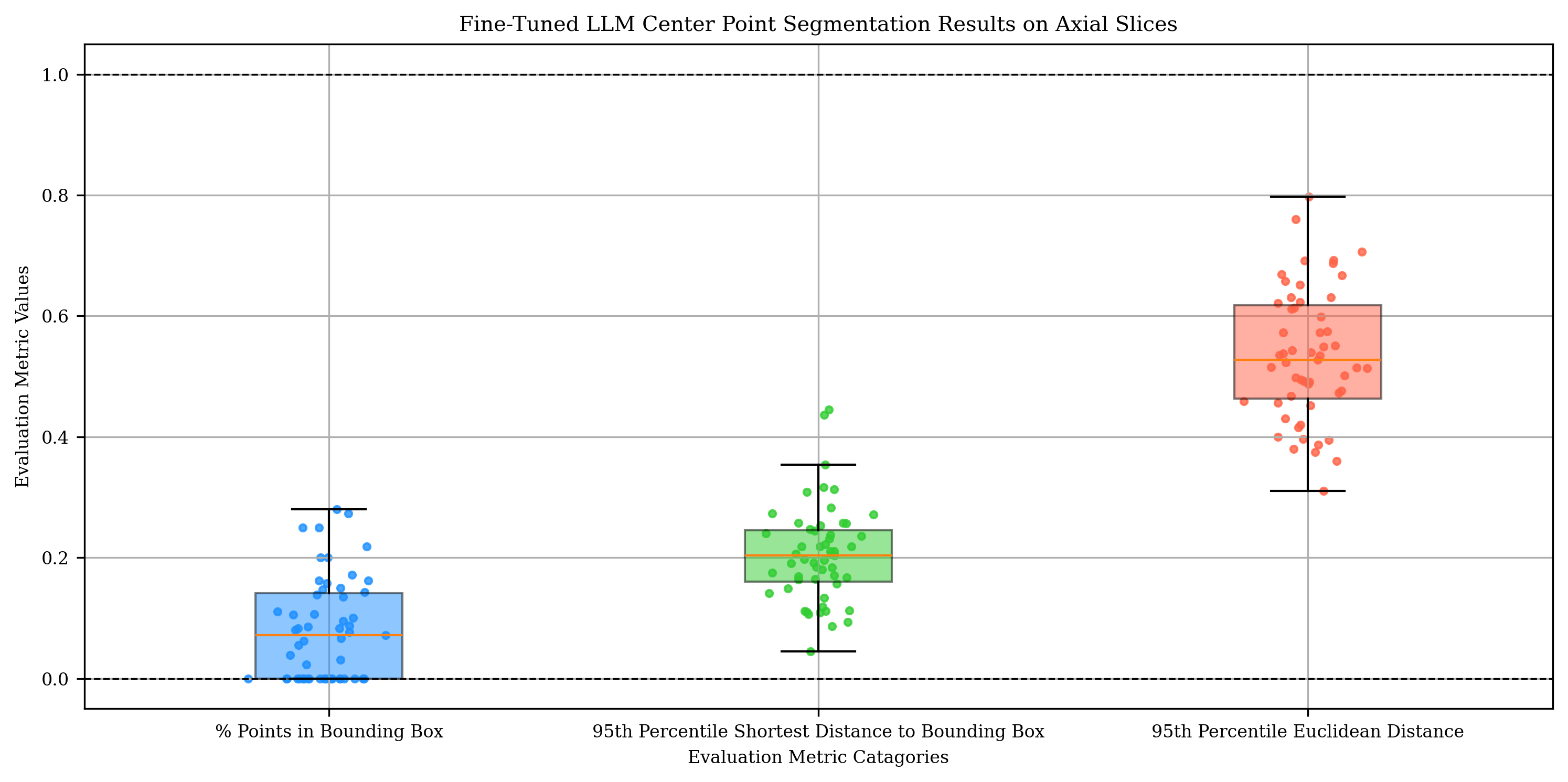}
\caption{Evaluation metrics for the fine-tuned LLM center point model predictions.
}
\label{fig_plotplot}
\end{figure}

The general center point model was fine-tuned for 100 steps with a batch size of 16. Figure~\ref{fig_plotplot} presents the evaluation results.

\begin{table}[h!]
\centering
\caption{Evaluation of the fine-tuned LLM center point model.}
\label{table33}
\begin{tabular}{lc}
\toprule
Metric & Value \\
\cmidrule(lr){1-1} \cmidrule(lr){2-2}
Average \% Points in Bounding Box & 0.0800 \\
Average 95\textsuperscript{th} Percentile Shortest Distance to Bounding Box & 26.2855 \\
Average 95\textsuperscript{th} Percentile Euclidean Distance & 68.4812 \\
\bottomrule
\end{tabular}
\end{table}

\begin{figure}[h!]
\centering
\includegraphics[width=\textwidth]{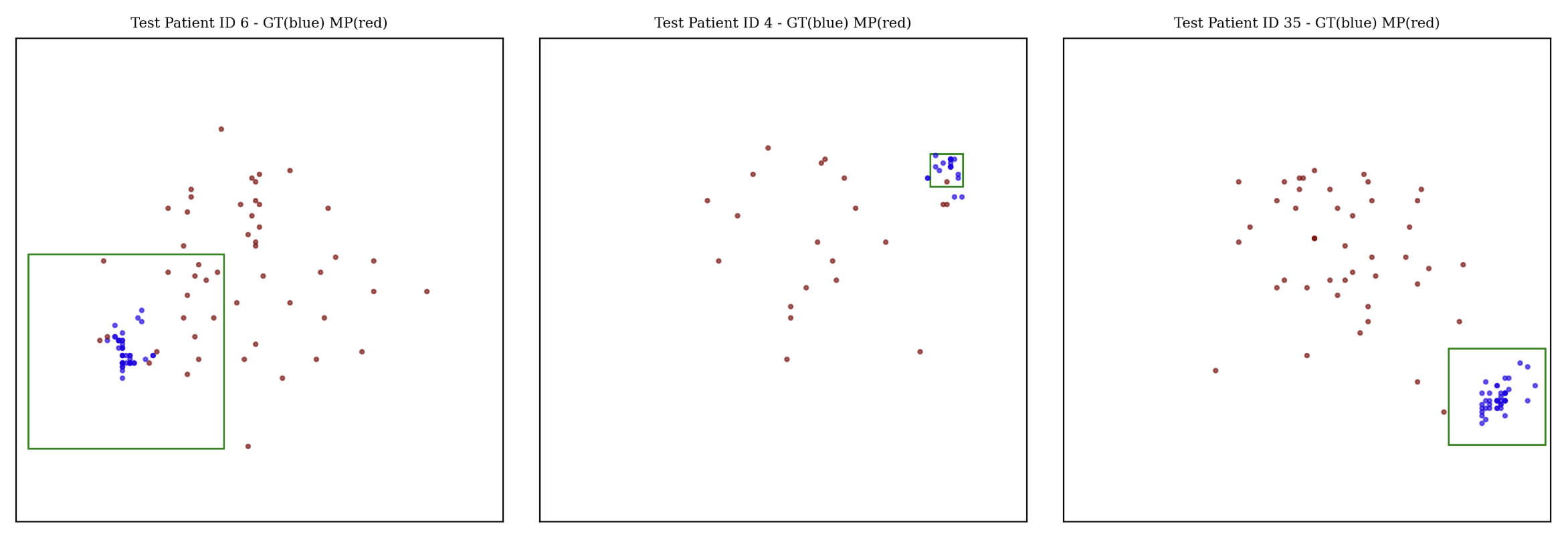}
\caption{Fine-tuned LLM center point segmentation visualization showed, for all patients, the model still predicted points near the center of the frame with more variance, regardless of the ground truth glioma location.
}
\label{fig_visvis}
\end{figure}

After fine-tuning, it was observed that the model's center point predictions became more spread out compared to the previous results, as shown in the visualizations in Figure~\ref{fig_visvis}. However, the points remained predominantly centered in the frame. In terms of distance metrics, the 95th percentile shortest distance to the bounding box (26.2855) and the 95th percentile Euclidean distance (68.4812) reflect the increased variance in predictions. As seen in Figure~\ref{fig_visvis}, some points are closer to the ground truth, while others are farther away, resulting in average distances that show no significant improvement compared to the initial model. 

In this segmentation task, the general LLM model's performance was assessed using the axial imaging orientation. The results are summarized in the table and Figure~\ref{fig_haha} below, providing an overview of the model's performance. 

\begin{figure}[h!]
\centering
\includegraphics[width=\textwidth]{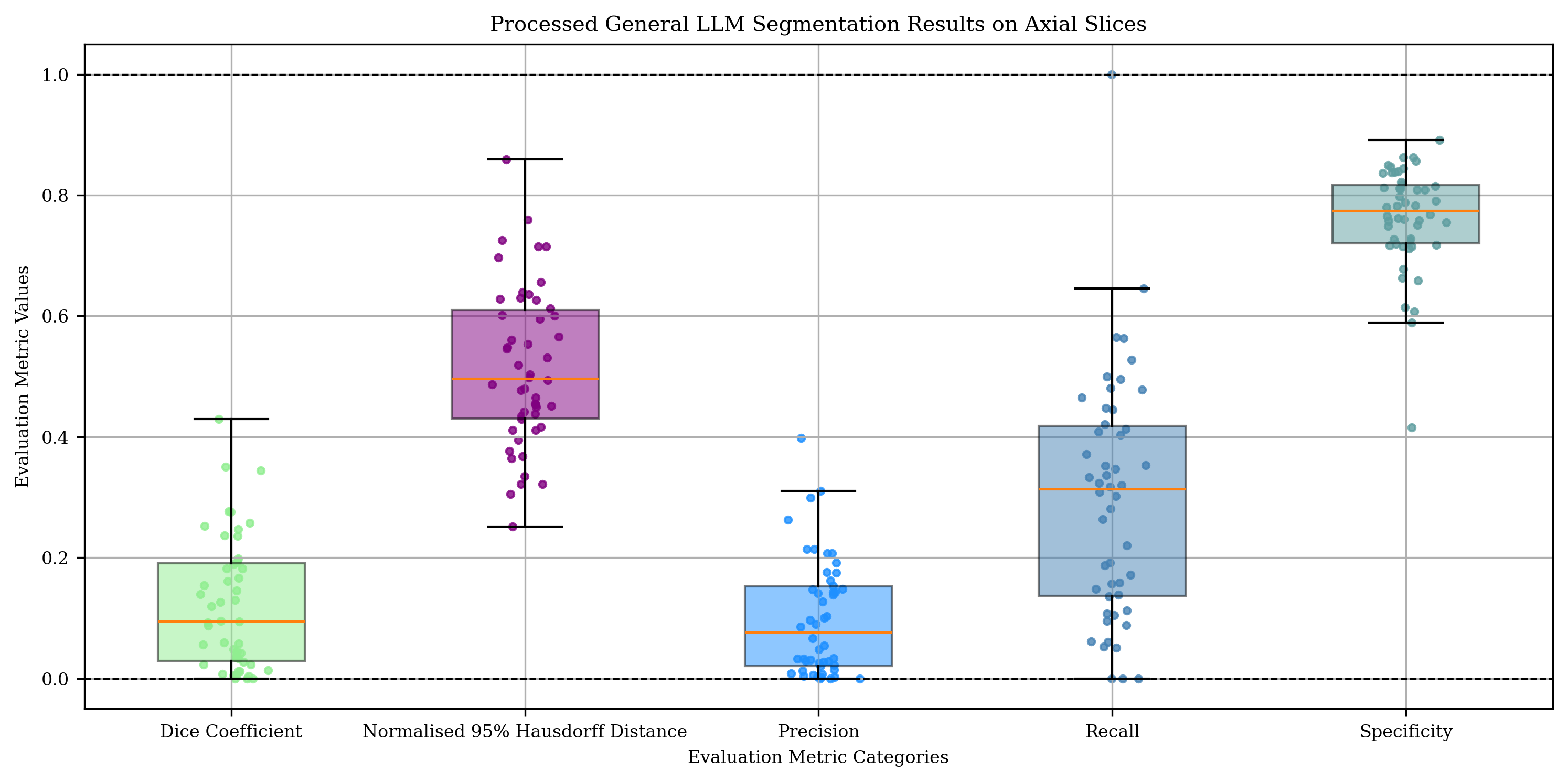}
\caption{Evaluation metrics for the general LLM bounding box model predictions.
}
\label{fig_haha}
\end{figure}

\begin{table}[h!]
\centering
\caption{Evaluation of the general LLM bounding box model.}
\label{table4}
\begin{tabular}{lc}
\toprule
Metric & Value \\
\cmidrule(lr){1-1} \cmidrule(lr){2-2}
Dice Coefficient & 0.1219 \\
95\textsuperscript{th} Hausdorff Distance & 65.9103 \\
Precision & 0.0989 \\
Recall & 0.2941 \\
Specificity & 0.7623 \\
\bottomrule
\end{tabular}
\end{table}

The general LLM model's performance on the bounding box test shows a Dice coefficient of 0.1219, indicating poor overlap between the predicted and actual tumor areas. The 95th percentile Hausdorff distance of 65.9103 suggests that the predicted segmentation is consistently far from the true tumor boundaries, reflecting significant localization errors. With a precision of 0.0989, only 9.89\% of the model's tumor predictions are correct, indicating a high rate of false positives and poor tumor identification. These results highlight the model's challenges in both accurately identifying and localizing gliomas.

The quantitative results presented above are further supported by visualizations of the ground truth and model-predicted bounding boxes. Examining these visualizations in the figures below shows several key observations regarding the general model's prediction behavior.

\begin{figure}[h!]
\centering
\includegraphics[width=\textwidth]{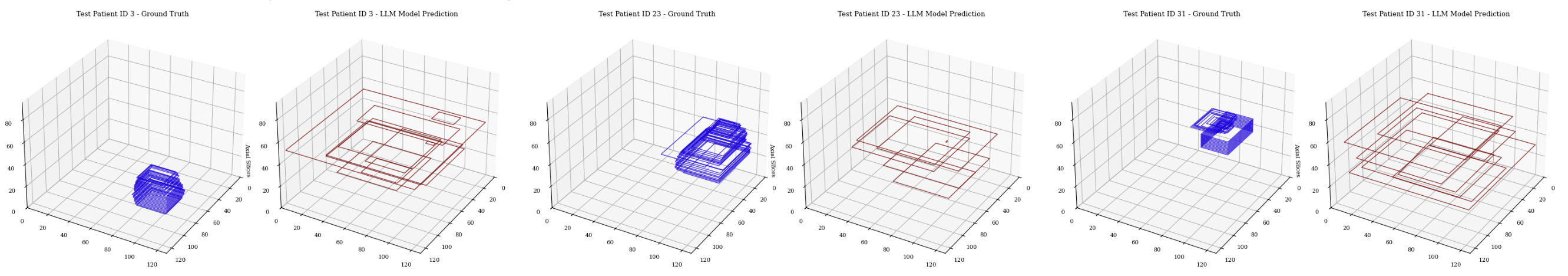}
\caption{General LLM bounding box segmentation visualization shows that the model's predictions are notably sparse.
}
\label{fig_sparse}
\end{figure}

\begin{figure}[h!]
\centering
\includegraphics[width=\textwidth]{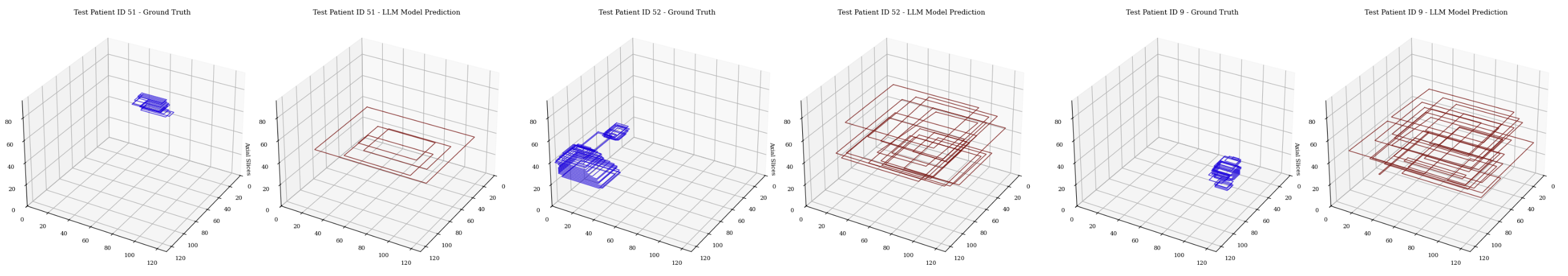}
\caption{General LLM bounding box segmentation visualization shows that the model consistently predicts gliomas near the center of the image.
}
\label{fig_locs}
\end{figure}

\begin{figure}[h!]
\centering
\includegraphics[width=\textwidth]{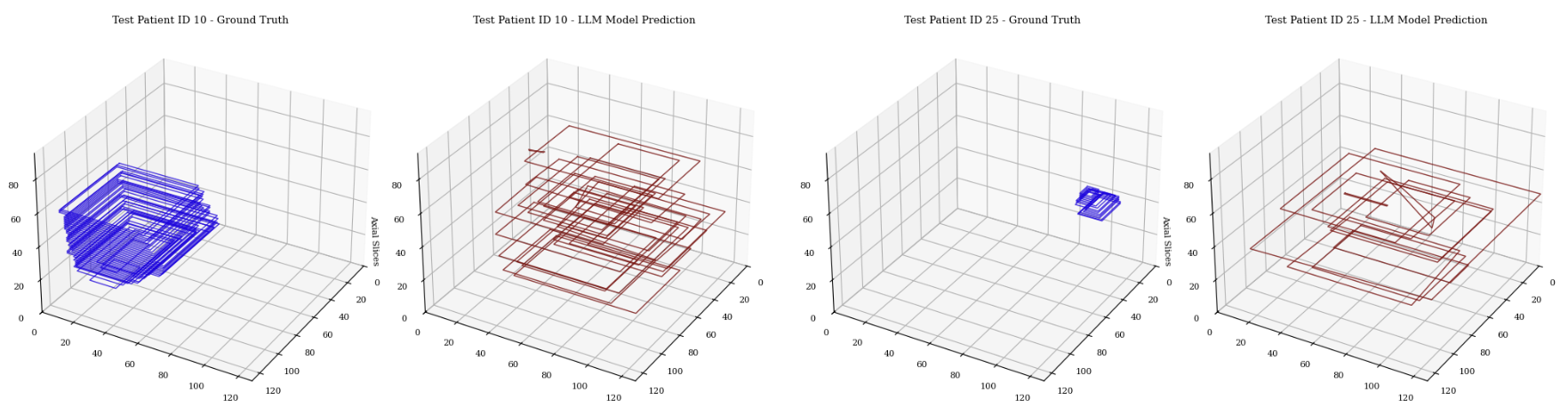}
\caption{General LLM bounding box segmentation visualization shows that the model's predictions lack clear differentiation between larger and smaller gliomas.
}
\label{fig_bigsmall}
\end{figure}

The model's predictions are notably sparse, as seen in Figure~\ref{fig_sparse}. In several instances, the LLM failed to respond in the expected format, often generating redundant explanations despite being provided with sufficient tokens. This resulted in fewer meaningful predictions, limiting the data available for analysis and making it challenging to obtain reliable results. We aimed to evaluate the general model’s ability to determine both the location and size of gliomas. In terms of location, as seen in Figure~\ref{fig_locs}, the model's predictions consistently fall in the middle of the image, regardless of the actual glioma positions. The predicted bounding boxes are frequently centered, showing little regard for the intensity or other relevant image features. This indicates that the model is not effectively utilizing the spatial information needed to accurately localize gliomas. When it comes to glioma size, as shown in Figure~\ref{fig_bigsmall}, the model’s predictions do not exhibit any clear bias between larger and smaller tumors. Regardless of the glioma’s actual size, the predicted bounding boxes remain similar in shape, with minimal variation and often concentrated in the middle of the frame. This suggests that the model struggles to capture differences in tumor size or spatial characteristics, further highlighting the limitations of the general LLM model in accurately segmenting gliomas. 

\begin{figure}[h!]
\centering
\includegraphics[width=\textwidth]{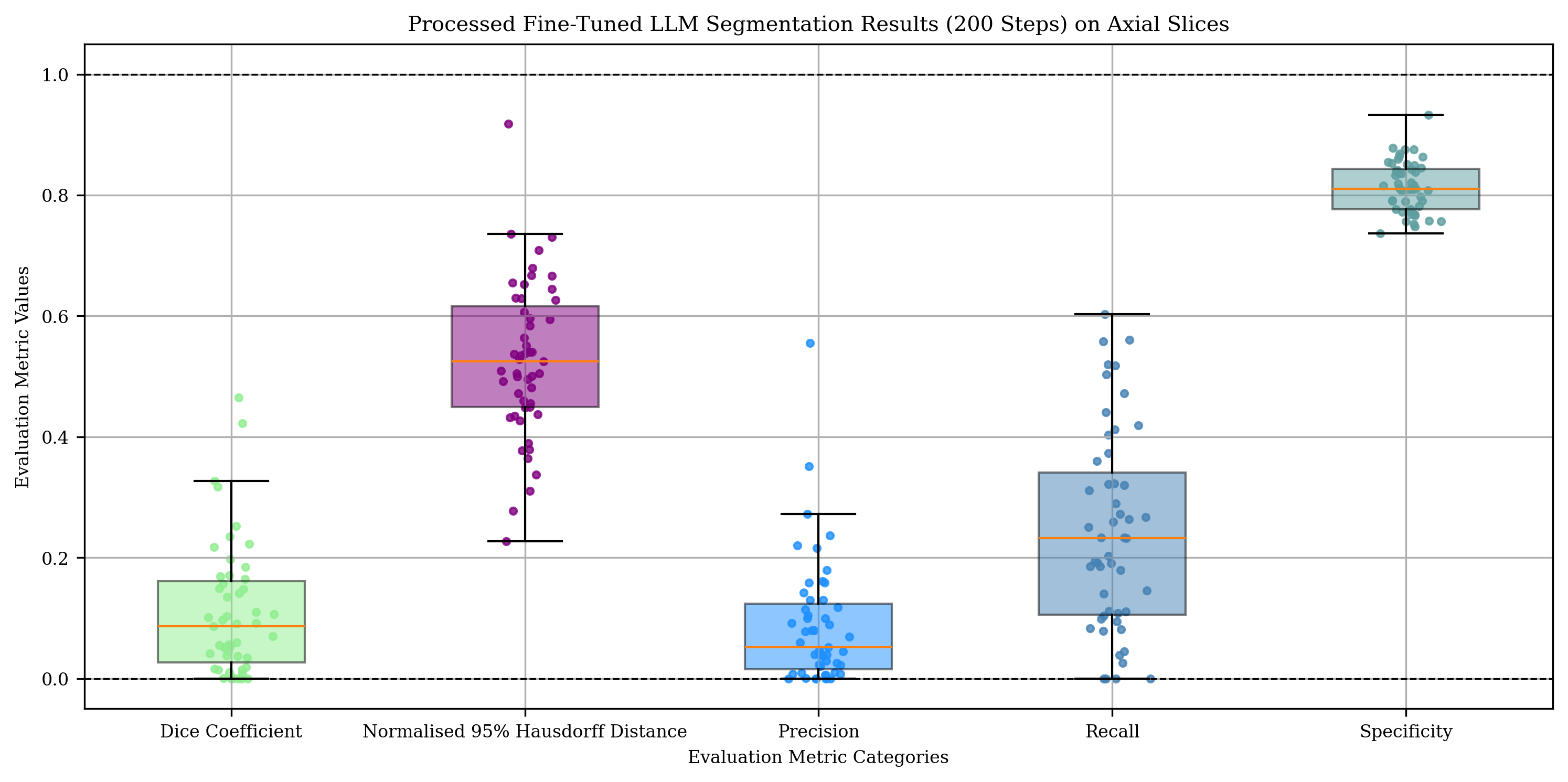}
\caption{Evaluation metrics for the fine-tuned LLM bounding box model predictions.
}
\label{fig_bbb}
\end{figure}

The general bounding box model was fine-tuned for 200 steps with a batch size of 16. Figure~\ref{fig_bbb} presents the evaluation results.

\begin{table}[h]
\centering
\caption{Evaluation of the fine-tuned LLM bounding box model.}
\label{table6}
\begin{tabular}{lc}
\toprule
Metric & Value \\
\cmidrule(lr){1-1} \cmidrule(lr){2-2}
Dice Coefficient & 0.1085 \\
95\textsuperscript{th} Hausdorff Distance & 67.3812 \\
Precision & 0.0882 \\
Recall & 0.2413 \\
Specificity & 0.8137 \\
\bottomrule
\end{tabular}
\end{table}

The Dice coefficient decreased to 0.1085, indicating that there was no significant improvement in the model's ability to segment the gliomas accurately. The 95\% Hausdorff distance slightly increased to 67.3812, suggesting that the model continues to struggle with aligning predicted center points with the true tumor locations. Precision decreased to 0.0882, while recall dropped to 0.2413, reflecting a decline in the model’s ability to correctly identify true positive centers. However, specificity improved to 0.8137, indicating that the model became more effective at minimizing false positives. Overall, the fine-tuning did not result in substantial improvements in the model’s performance.

\begin{figure}[h!]
\centering
\includegraphics[width=\textwidth]{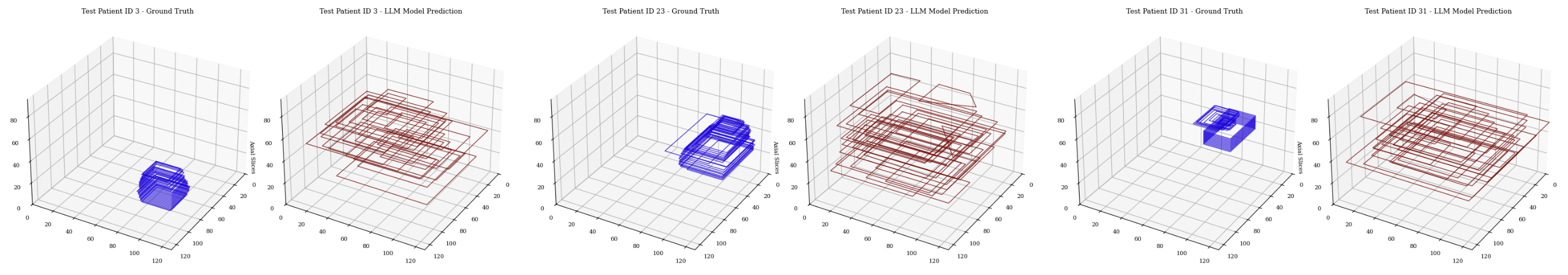}
\caption{Fine-tuned LLM bounding box segmentation visualization shows no noticeable improvement in the placement of bounding boxes.
}
\label{fig_qw}
\end{figure}

\begin{figure}[h!]
\centering
\includegraphics[width=\textwidth]{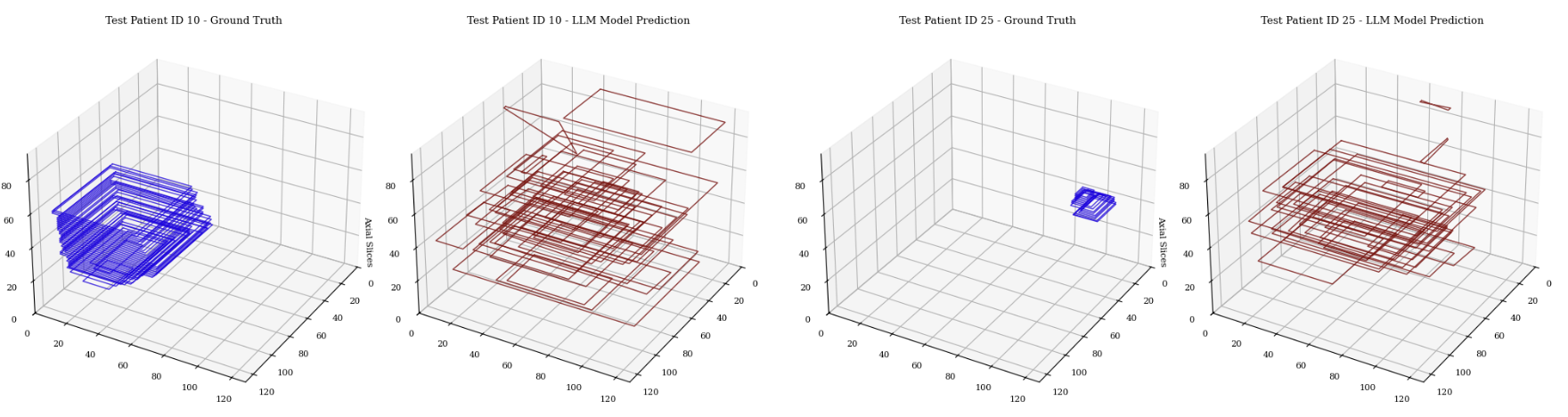}
\caption{Fine-tuned LLM bounding box segmentation visualization shows continued insensitivity to glioma size.
}
\label{fig_qe}
\end{figure}

The qualitative visualizations of the bounding box segmentations show some improvement in the model's ability to generate responses, as seen in Figure~\ref{fig_qw} and Figure~\ref{fig_qe}. Previously, the model struggled to provide consistent answers, but now it regularly produces more bounding box predictions across slices. However, despite the increase in responses, there is no noticeable improvement in the placement of the bounding boxes after fine-tuning. The bounding boxes remain in similar locations as before, as shown in Figure~\ref{fig_qw}, and the model continues to show no sensitivity to the size of the gliomas, as seen in Figure~\ref{fig_qe}. This suggests that the model is not adjusting its predictions based on tumor size. It still tends to place bounding boxes near the center of the image, often on the larger side. While this may minimize Dice loss, it indicates that the model has not learned to segment gliomas effectively. 

In this last segmentation task, the general LLM model's performance was evaluated using the bounding polygon test again with just the axial imaging orientation. The results, as summarized in the table and Figure~\ref{fig_hehe} below, providing an overview of the model’s performance.

\begin{figure}[h!]
\centering
\includegraphics[width=\textwidth]{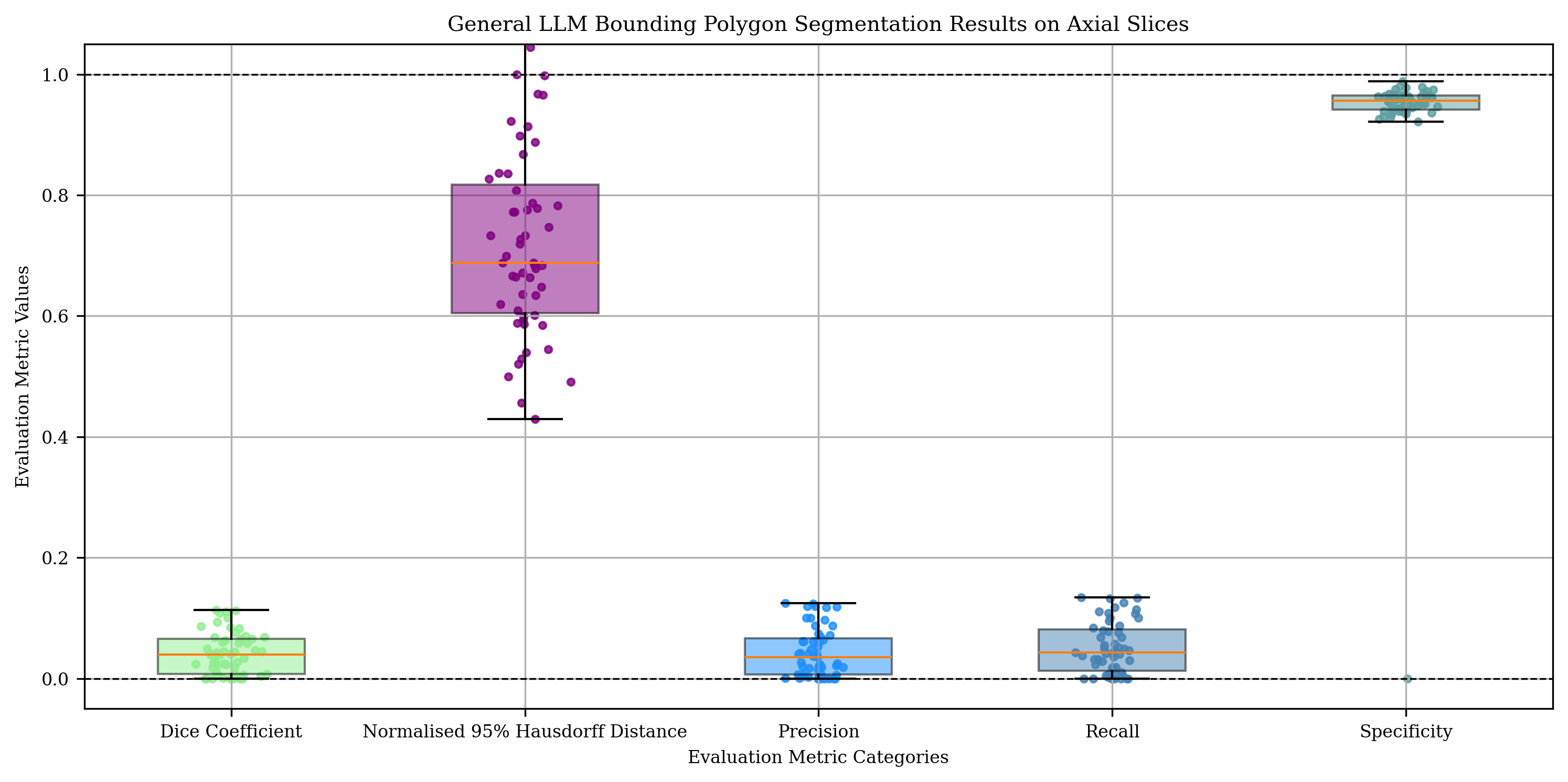}
\caption{Evaluation metrics for the general LLM bounding polygon model predictions.
}
\label{fig_hehe}
\end{figure}

\begin{table}[h]
\centering
\caption{Evaluation of the general LLM bounding polygon model.}
\label{table7}
\begin{tabular}{lc}
\toprule
Metric & Value \\
\cmidrule(lr){1-1} \cmidrule(lr){2-2}
Average Dice Coefficient & 0.0412 \\
Average 95\% Hausdorff Distance & 92.4296 \\
Average Precision & 0.0432 \\
Average Recall & 0.0502 \\
Average Specificity & 0.9371 \\
\bottomrule
\end{tabular}
\end{table}

The general LLM model's performance on the bounding polygon test shows a Dice coefficient of 0.0412, indicating a very low overlap between the predicted and actual tumor areas. The 95th percentile Hausdorff distance of 92.4296 suggests that the predicted segmentation is significantly distant from the true tumor boundaries, reflecting substantial localization errors. With a precision of 0.0432, only 4.32\% of the model's tumor predictions are correct, suggesting a high rate of false positives and poor tumor identification. Recall is also low at 0.0502, indicating that the model is not effectively identifying a significant proportion of the actual tumors. However, the specificity of 0.9371 shows that the model is good at avoiding false positives, correctly identifying regions of the image where tumors are not present.

\begin{figure}[h!]
\centering
\includegraphics[width=\textwidth]{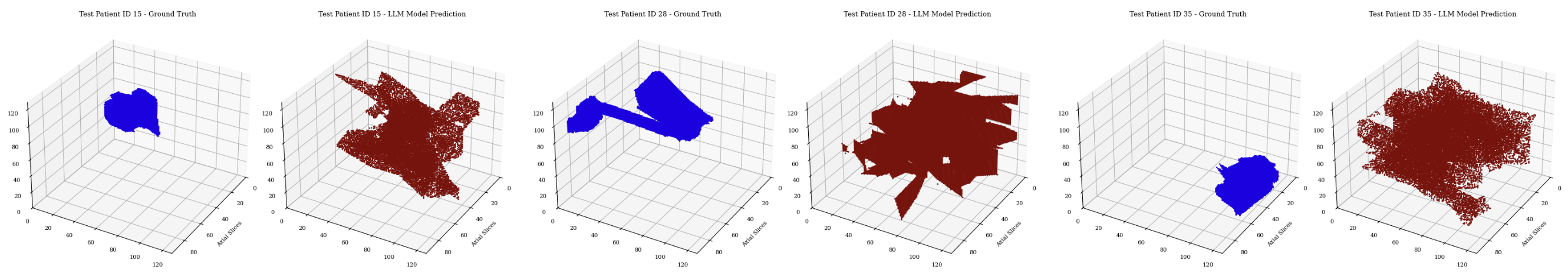}
\caption{General LLM bounding polygon segmentation visualization showed random predicted shapes, centered in the image with varying sizes.
}
\label{fig_xxx}
\end{figure}

The visualizations of the bounding polygon segmentations did not reveal any discernible trends or improvements. As seen in Figure~\ref{fig_xxx}, the predicted shapes appeared random, varying in size and consistently placed in the center of the image. This shows no clear ability to localize tumors to specific ground truth sizes or locations. For adjacent slices, which are often very similar in scans and would logically require similar segmentations, the model made drastically different predictions for each slice. As a result, the polygon segmentations resembled irregular, star-like shapes, emphasizing the model’s inability to localize and segment gliomas across slices. 

\begin{figure}[h!]
\centering
\includegraphics[width=\textwidth]{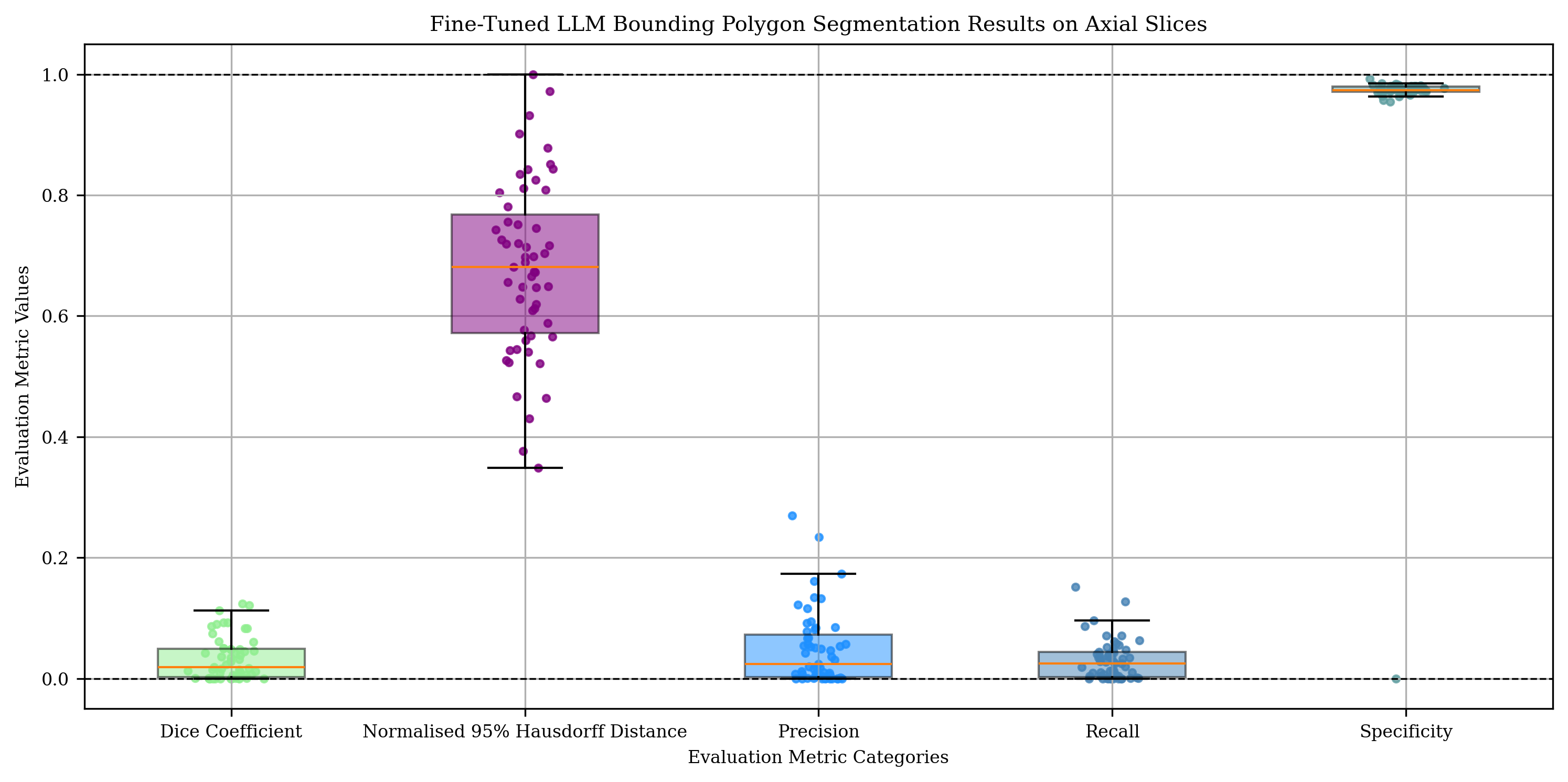}
\caption{Evaluation metrics for the fine-tuned LLM bounding polygon predictions.
}
\label{fig_ooo}
\end{figure}

The general bounding polygon model was fine-tuned for 150 steps with a batch size of 4. Figure~\ref{fig_ooo} presents the evaluation results.

\begin{table}[h]
\centering
\caption{Evaluation of the fine-tuned LLM bounding polygon model.}
\label{table8}
\begin{tabular}{lc}
\toprule
Metric & Value \\
\cmidrule(lr){1-1} \cmidrule(lr){2-2}
Average Dice Coefficient & 0.0335 \\
Average 95\% Hausdorff Distance & 86.9709 \\
Average Precision & 0.0487 \\
Average Recall & 0.0296 \\
Average Specificity & 0.9568 \\
\bottomrule
\end{tabular}
\end{table}

The Dice coefficient decreased to 0.0335, indicating that the fine-tuning did not significantly enhance the model’s ability to segment the gliomas accurately. The 95\% Hausdorff distance slightly improved to 86.9709, suggesting that while the model is still struggling with the precise alignment of predicted tumor boundaries, it has not shown substantial progress. Precision increased to 0.0487, though still low, indicating a slight improvement in the model’s ability to correctly identify true tumor boundaries. However, recall dropped to 0.0296, showing that the model is still failing to capture a significant portion of the actual tumors. Specificity improved to 0.9568, demonstrating that the model has become more effective at avoiding false positives. Overall, the fine-tuning resulted in marginal improvements but did not lead to significant performance gains in accurately segmenting gliomas.

\begin{figure}[h!]
\centering
\includegraphics[width=\textwidth]{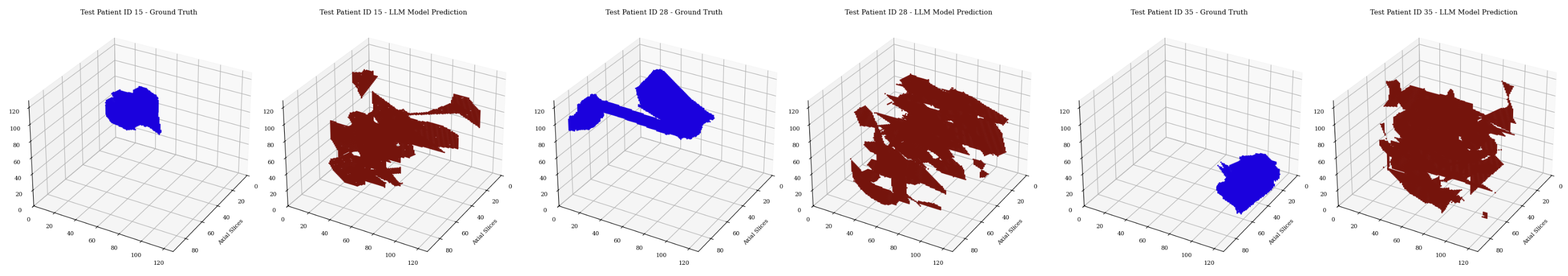}
\caption{Fine-tuned LLM bounding polygon segmentation visualization shows that the segmentations remain largely unchanged. 
}
\label{fig_mmm}
\end{figure}

After fine-tuning, the bounding polygon predictions appeared slightly more concentrated around the center of the image, as shown in Figure~\ref{fig_mmm}, but overall, they remain largely unchanged from the previous results. This suggests that fine-tuning did not lead to significant improvements in the model's ability to localize or accurately segment gliomas. The pattern of random, inconsistent shapes persists, and the segmentations still did not align with the actual tumor locations or sizes. This raises the possibility that the current evaluation method may not have been the most effective in capturing meaningful improvements in the model's performance. 

\section{Discussion}
\subsection{Key Findings}

\subsubsection{Image classification}
The core objective of this research was to assess and compare the performance of general-purpose large language models (LLMs) and subspecialized models in medical imaging tasks, particularly focusing on their accuracy, robustness, and practical utility in data-limited image-based environments. The evaluation centers on determining when fine-tuning general LLM models provides measurable benefits in performance over simply using them for inference out-of-the-box.

The CNN baseline demonstrated competitive performance with a training accuracy of 0.9677, validation accuracy of 0.7581, and testing accuracy of 0.8000. Precision and recall were particularly strong in the training phase, achieving 1.0000 and 0.9592, respectively, while the testing phase maintained a balanced F1 score of 0.8706. The AUC remained high in testing (0.8202), although specificity dropped to 0.6364, indicating some difficulty in correctly identifying negative cases (LGG). The baseline as a whole shows it is accurate and robust.

In contrast, the general LLM model showed promising accuracy and recall across axial, coronal, and sagittal orientations, with the coronal view achieving the highest accuracy (0.8000) and perfect recall (1.0000). However, these metrics were misleading due to the dataset's class imbalance, where high-grade glioma (HGG) slices vastly outnumbered low-grade glioma (LGG) slices. The model predominantly predicted HGG for nearly all samples, resulting in near-zero specificity across orientations. The GLDR visualizations further emphasized this issue, as most patient scans had HGG favouring ratios, even in cases where LGG was the ground truth. The expectation was to observe more balanced predictions in borderline cases due to the "winner-takes-all" voting system, but this was not reflected in the results. In its current out-of-the-box form, the general LLM model’s inability to distinguish between HGG and LGG limits its clinical utility and classification reliability. 

The results for the full-epoch and 100-step fine-tuned models show limited improvement over the general model. The full-epoch model showed a decrease in performance, with lower accuracy (0.6667) and recall (0.7073) compared to the 100-step model, which maintained a higher recall of 0.875. Neither model demonstrated significant improvements in distinguishing between HGG and LGG, as the predictions remained split between the two types, contrary to expectations of more distinct clustering. This suggests that the fine-tuning process has not yet effectively enabled the model to differentiate between glioma types, particularly LGG gliomas, with both models showing only marginal differences. The results indicate that further refinement is necessary to achieve more accurate classification.

These findings align with the study's hypothesis that general-purpose LLMs, while versatile and successfully applicable into a variety of medical tasks, are not inherently well-suited for specialized tasks like medical image classification even with fine-tuning. The CNN’s stronger performance in differentiating glioma types shows the advantage of domain-specific models in handling class imbalance and nuanced medical data. 

\subsubsection{Image segmentation}
The core objective of this research was to evaluate and compare the performance of general-purpose large language models (LLMs) and subspecialized models in medical image segmentation tasks, with a focus on their accuracy, robustness, and practical utility in environments with limited labeled data. The evaluation aimed to assess when fine-tuning general LLMs provides measurable benefits over simply using them for inference out-of-the-box in the context of segmentation.

The CNN baseline model showed solid performance in glioma segmentation, achieving a Dice coefficient of 0.5942 on the testing set. It successfully segmented larger gliomas, accurately capturing tumor boundaries and general shapes. In these cases, the model aligned well with the ground truth, demonstrating reliable tumor detection and boundary delineation. However, challenges emerged with smaller gliomas and those with irregular shapes. For these tumors, the model struggled to accurately localize and define boundaries, sometimes predicting scattered points instead of a well-defined tumor shape. This was particularly noticeable with gliomas of unusual sizes or strange shapes, where the model tended to either overestimate or under-represent the tumor extent. Additionally, the model occasionally over-segmented the brain, capturing broader regions beyond the glioma, such as other brain structures or anomalies. This over-segmentation could have been due to low contrast in the images or the presence of other brain features that the model misidentified as part of the tumor. Despite this, the model still demonstrated a strong ability to identify and segment gliomas in more straightforward cases, with fewer issues in detecting larger or more distinct tumors.

The general LLM model's performance in center point segmentation was notably poor, with the predicted points mostly scattered around the center of the image rather than aligning with the true tumor location. Only 8.05\% of the predicted points fell within the ground truth bounding box, reflecting a random approach to identifying tumor centers. The high 95th percentile Euclidean and shortest distances (57.4112 and 22.8566, respectively) further demonstrated that the model struggled to accurately localize gliomas, often predicting points far from the actual tumor locations. After fine-tuning, there was a slight increase in the variance of the predictions, indicating the model was attempting to explore different regions of the image. However, the improvement was minimal, with the percentage of points within the bounding box remaining nearly unchanged at 8.00\%. The distance metrics slightly worsened (95th percentile Euclidean distance: 68.4812, shortest distance: 26.2855), suggesting that the model had not learned to focus on identifying the correct tumor centers. Instead, the predictions remained centered in the image. These results highlight that fine-tuning did not yield meaningful improvements in the model's ability to localize gliomas, and the predictions remained largely random.

The general LLM model's performance on the bounding box segmentation task also revealed significant shortcomings. With a Dice coefficient of 0.1219, the model showed poor overlap between the predicted and actual tumor regions. The 95th percentile Hausdorff distance of 65.9103 indicated that the predicted tumor boundaries were consistently far from the true tumor locations, and with a precision of 0.0989 and recall of 0.2941, the model struggled to accurately identify and localize gliomas, with a high rate of false positives and missed tumor regions. Visualizations revealed that the model's bounding boxes were often centered in the image, irrespective of the tumor's actual location or size, suggesting that the model did not effectively utilize the spatial information or tumor features, highlighting its limitations in precise tumor segmentation. After fine-tuning for 200 steps, the model's performance showed little improvement. The Dice coefficient slightly decreased to 0.1085, and the 95th percentile Hausdorff distance rose to 67.3812, reflecting the continued difficulty in aligning predicted bounding boxes with actual tumor boundaries. Precision and recall further declined to 0.0882 and 0.2413, respectively, indicating a deterioration in the model's ability to detect true positive centers. However, specificity improved to 0.8137, suggesting better handling of false positives. Despite these marginal changes, qualitative analysis of the bounding box segmentations revealed that while the model was now consistently producing more bounding box predictions, the placement of these boxes remained largely unchanged, and were still placed near the center of the image. The model also showed no sensitivity to the gliomas' size, indicating it had not learned to make the critical distinctions necessary for accurate segmentation. Instead of focusing on relevant tumor features, the model's predictions remained centered and arbitrary, demonstrating that fine-tuning did not lead to meaningful improvements in performance.

The general LLM model's performance on the bounding polygon segmentation task also yielded poor results. The Dice coefficient of 0.0412 indicates minimal overlap between predicted and true tumor areas, while a 95th percentile Hausdorff distance of 92.4296 highlights significant localization errors. Precision (0.0432) and recall (0.0502) were very low, reflecting a high false-positive rate and poor tumor identification. However, specificity was high (0.9371), indicating the model avoided false positives but failed to accurately segment tumors. Visualizations showed random, irregular shapes centered in the image, with no correlation to actual tumor locations or sizes. Adjacent slices, which should have similar segmentations, showed drastically different predictions. After fine-tuning for 150 steps, the model showed no significant improvements, with the Dice coefficient decreasing to 0.0335, the Hausdorff distance improving slightly to 86.9709, precision rising to 0.0487, and specificity increasing to 0.9568. However, recall dropped to 0.0296, and the bounding polygons still failed to align with true tumor locations, suggesting limited impact from fine-tuning and ineffective evaluation.

The findings of this study align with the hypothesis that general-purpose LLMs, despite their versatility in a range of medical tasks, are not inherently suited for specialized tasks like medical image segmentation, even with fine-tuning.

\subsection{Limitations}

\subsubsection{Image classification}
A major limitation of this study is the difference in input data between the CNN and the LLM models. The CNN used full 3D convolutions and all four MRI modalities (FLAIR, T1, T1ce, and T2), giving it access to rich spatial and multi-modal information. This comprehensive input may contribute to its stronger performance. In contrast, the LLM was restricted to 2D axial slices from only the FLAIR modality given its necessary input structure for prompting. Each slice was classified individually, and patient-level predictions were determined by majority vote across slices. This approach limited the LLM’s access to complete spatial and modality information. Although it’s important to acknowledge this difference when evaluating the methods and understanding how the comparison was conducted, it ultimately has minimal impact on the study’s goals. The primary focus is on comparing the best-performing versions of the CNN and LLM to assess their overall performance, regardless of input disparities.

Another key limitation is the inconsistency in the general-purpose LLM’s predictions. In consistency testing, the LLM correctly identified an HGG in all 95 trials. However, for an LGG, it misclassified the case as HGG 77 times and correctly identified it only 18 times. This inconsistency, despite identical inputs, highlights the model’s instability and unreliability, which also translate into uncertainty in the investigation of the general LLM. We saw further in this consistency test that fine-tuning failed to achieve better results. Instead of improving the model's ability to distinguish between HGG and LGG, the fine-tuning process appears to have introduced additional instability. The fact that the model's predictions for LGG images are nearly evenly split between LGG and HGG classifications suggests that the model has not developed a clear decision boundary between the two classes.

The next significant limitation in this study is the restricted scope of fine-tuning performed on the LLM. The fine-tuning process could have been far more robust, but was heavily limited by resource constraints. The batch size of just 16 2D images per iteration was quite small, especially when compared to the CNN, which used 16 3D patients with four MRI modalities per sample. This was due to the memory constraints of Colab GPUs. The batch size can heavily impact the model's ability to generalize, potentially hindering the LLM’s capacity to capture the complex patterns needed to differentiate between glioma subtypes with fine-tuning. In addition to the batch size, the fine-tuning process was constrained by the relatively short training duration. With only 100 steps and one full epoch completed in these fine-tuned models, they may not have had enough time to converge to an optimal solution. While the training ran for over 10 hours, the limited number of iterations meant the LLM was not exposed to enough data to learn robust features and exploit the underlying patterns in the data, which could have been better captured with additional epochs. The extended runtime was costly on Colab, and further training would have incurred significant expenses, limiting the ability to run additional epochs. Another challenge during fine-tuning was the absence of a validation curve. Unlike the CNN training, which included regular validation checks to monitor for overfitting, the fine-tuning of the LLM did not include such checks due to the extensive time constraints. Without this important feedback, it was difficult to assess whether the model was overfitting to the training data or if the fine-tuning was progressing effectively. To mitigate this, very small learning rates were used in an attempt to make incremental updates, but without the ability to validate the model’s performance consistently, there was uncertainty regarding how well the model was learning. The combination of these resource limitations, small batch size, limited training duration, and lack of validation, likely contributed to suboptimal fine-tuning results for both subspecialised LLM models.

\subsubsection{Image segmentation}
A major limitation of this study is the difference in input data between the CNN and the LLM models, which mirrors the limitation observed in the classification task. The CNN utilized full 3D convolutions and all four MRI modalities (FLAIR, T1, T1ce, and T2), providing rich spatial and multi-modal information, which likely contributed to its stronger performance in segmentation. In contrast, the LLM was restricted to 2D axial slices from only the FLAIR modality. Each slice was processed independently, with patient-level predictions determined by majority voting across slices. This restricted input, while necessary for the LLM’s prompting structure, limited its access to full spatial and modality information, affecting its segmentation performance.

The restricted scope of fine-tuning, was another limitation shared with the classification task, also hindered the LLM’s performance in segmentation. The small batch size of 16 2D images per iteration, limited by Colab GPU memory constraints, and the short training duration of just 100 to 200 steps, significantly impacted the model’s ability to learn robust features. This was compounded by the lack of validation checks, which would have allowed for better monitoring of the model’s learning process. While small learning rates were used to mitigate overfitting, the absence of a validation curve introduced uncertainty regarding the model’s true performance. These resource constraints likely contributed to the suboptimal fine-tuning results, which were consistent with the issues encountered during classification.

Another limitation was the different prompting strategies required to enable the LLM to perform segmentation. Unlike the CNN, which directly learns spatial relationships from voxel-based 3D data, the LLM requires text-based input for its segmentation task. This conversion of voxel locations into tokens for prompting could have introduced complexity in learning spatial relationships, as the model was tasked with interpreting segmentation as a sequence of tokens rather than directly processing physical voxel locations. Converting these locations back into segmentation predictions likely added an additional layer of abstraction, making it more difficult for the model to learn accurate tumor boundaries. This approach could have impeded the model's ability to effectively capture spatial details and localize gliomas, which are typically better represented by the direct manipulation of voxel-based data in models like CNNs.

\section{Conclusion}

For the image classification task, the CNN model outperformed the general-purpose LLM in terms of accuracy, robustness, and reliability in classifying glioma types, particularly in handling class imbalances and distinguishing between LGG and HGG gliomas. While the LLM showed some promise, its initial performance was hindered by low specificity and inconsistent predictions, especially for LGG cases. Although fine-tuning was attempted, the results were not as effective as anticipated due to limitations in training duration, batch size, and resource constraints. The LLM’s potential remains, but further refinement and more robust fine-tuning are necessary for it to effectively tackle specialized medical imaging tasks like glioma classification.

For the image segmentation task, the CNN model outperformed the general-purpose LLM in terms of accuracy, robustness, and reliability in segmenting gliomas, particularly in accurately identifying tumor boundaries and handling varying tumor sizes. While the LLM showed some potential, its initial performance was limited by poor spatial localization and high variability in segmentation results. Fine-tuning the LLM did not yield significant improvements due to constraints in training duration, batch size, and available resources. Although the LLM has potential for medical image segmentation, further refinement, more robust fine-tuning, and improvements in input processing are needed for it to effectively handle specialized tasks like glioma segmentation.

In conclusion, this research provided valuable insights into the effectiveness of general-purpose large language models (LLMs) versus specialized models in medical imaging tasks, particularly in glioma classification and segmentation. Our findings demonstrated that while general-purpose LLMs showed some promise, they were significantly outperformed by specialized models, such as Convolutional Neural Networks (CNNs), in terms of accuracy, robustness, and handling class imbalances. The CNN baseline excelled in both glioma classification and segmentation, achieving strong performance metrics and effectively addressing challenges like class imbalance and localization errors. In contrast, the LLM struggled, with issues such as low specificity, inconsistent predictions, and difficulties in learning spatial relationships due to its text-based nature. Fine-tuning the LLM resulted in only marginal improvements, hindered by constraints like limited data, small batch sizes, and insufficient training duration.

These findings suggest that while general-purpose LLMs have potential, they are not yet well-suited for specialized medical tasks like glioma classification and image segmentation. Specialized models, particularly CNNs, proved to be more accurate and robust, underscoring the importance of domain-specific models in handling complex medical data. This research emphasizes the need for tailored approaches in healthcare AI, offering a clearer understanding of when and where LLMs are most effective and establishing a framework for selecting optimal models and training strategies for various medical tasks. Ultimately, fine-tuning was found to have limited impact unless substantial adaptations were made to align the models more closely with the specific requirements of the medical tasks at hand.

There is still more work to be done in this area. First, more rigorous fine-tuning with larger batch sizes, longer training durations, and proper validation will be necessary. This will allow for a more robust comparison of the fine-tuned results, ensuring that the improvements seen are genuine and not influenced by resource constraints or training limitations. Additionally, it will be crucial to determine whether these findings are replicable across different datasets. While this study primarily focused on the BraTS dataset, it would be valuable to explore whether similar trends emerge when applied to other datasets. This could help assess the generalizability of our results and identify any dataset-specific differences or challenges. Evaluating the value of text-based LLMs in image-based tasks raises the immediate question of how these models compare to foundational models dedicated to imaging tasks. 

A further comparative study between text-based LLMs and specialized foundational imaging models would also provide insights into their relative performance and highlight strengths and weaknesses in different contexts.


\bibliography{refs}


\end{document}